%% file: main.tex
\documentclass[10pt,twocolumn,letterpaper]{article}

\usepackage[pagenumbers]{cvpr} % To force page numbers, e.g. for an arXiv version

\input{preamble}
\usepackage{marvosym}
\makeatletter
\def\customsymbol#1{
    \ifcase\number\value{#1}
        \or \Letter
    \else\@ctrerr
    \fi
}
\makeatother
\definecolor{cvprblue}{rgb}{0.21,0.49,0.74}
\usepackage[pagebackref,breaklinks,colorlinks,allcolors=cvprblue]{hyperref}

\def\paperID{20259} % *** Enter the Paper ID here
\def\confName{CVPR}
\def\confYear{2026}

\title{Learning 3D Representations for Spatial Intelligence \\ from Unposed Multi-View Images}

\author{Bo Zhou$^{1}$,~~\hspace{1pt}Qiuxia Lai$^{3}$,~~\hspace{1pt}Zeren Sun$^{1}$,~~\hspace{1pt}Xiangbo Shu$^{1}$,~~~\hspace{1pt}Yazhou Yao$^{1}$,~~\hspace{1pt}Wenguan Wang$^{2}$\textsuperscript{\Letter}   \\
    \small {$^1$} Nanjing University of Science and Technology,
    \small {$^2$}  Zhejiang University,
    \small {$^3$}  Communication University of China,  \\
    \small \url{https://bobochow.github.io/UniSplat}
}

\begin{document}
\begingroup
\maketitle
\endgroup
\footnotetext[1]{Corresponding author.}
\input{sec/0_abstract}    
\input{sec/1_intro}
\input{sec/2_related}

\input{sec/3_method}

\input{sec/4_experiments}

\input{sec/5_conclusion}
{
    \small
    \bibliographystyle{ieeenat_fullname}
    \bibliography{main,supp}
}
\input{sec/X_suppl}

\end{document}

%% file: preamble.tex
\usepackage{url}
\usepackage{graphicx}
\usepackage{bbm}
\usepackage{colortbl}
\usepackage{multirow}
\usepackage{enumitem}
\usepackage{float}
\usepackage{placeins}
\usepackage{tabularx}
\usepackage{wrapfig}
\usepackage{caption}
\usepackage{makecell}
\usepackage{vcell}
\makeatletter
\newcommand{\thickhline}{%
	\noalign {\ifnum 0=`}\fi \hrule height 1pt
	\futurelet \reserved@a \@xhline
}
\makeatother
\definecolor{mygray}{gray}{.9}
\usepackage{stfloats}
\usepackage{bm}

%% file: sec/0_abstract.tex
\begin{abstract}
Robust 3D representation learning forms the perceptual foundation of spatial intelligence, enabling downstream tasks in scene understanding and embodied AI. However, learning such representations directly from unposed multi-view images remains challenging. Recent self-supervised methods attempt to unify geometry, appearance, and semantics in a feed-forward manner, but they often suffer from weak geometry induction, limited appearance detail, and inconsistencies between geometry and semantics.
We introduce \textbf{\textit{UniSplat}}, a feed-forward framework designed to address these limitations through three complementary components. 
First, we propose a \textit{dual-masking strategy} that strengthens geometry induction in the encoder. By masking both encoder and decoder tokens, and targeting decoder masks toward geometry-rich regions, the model is forced to infer structural information from incomplete visual cues, yielding geometry-aware representations even under unposed inputs.
Second, we develop a \textit{coarse-to-fine Gaussian splatting strategy} that reduces appearance-semantics inconsistencies by progressively refining the radiance field.
Finally, to enforce geometric–semantic consistency, we introduce a \textit{pose-conditioned recalibration mechanism} that interrelates the outputs of multiple heads by reprojecting predicted 3D point and semantic maps into the image plane using estimated camera parameters, and aligning them with corresponding RGB and semantic predictions to ensure cross-task consistency, thereby resolving geometry–semantic mismatches. 
Together, these components yield unified 3D representations that are robust to unposed, sparse-view inputs and generalize across diverse tasks, laying a perceptual foundation for spatial intelligence.
\end{abstract}

%% file: sec/1_intro.tex
\section{Introduction}
\label{sec:intro}

\begin{figure}[!ht]
\setlength{\abovecaptionskip}{2pt}
\centering
\includegraphics[width=0.95\linewidth]{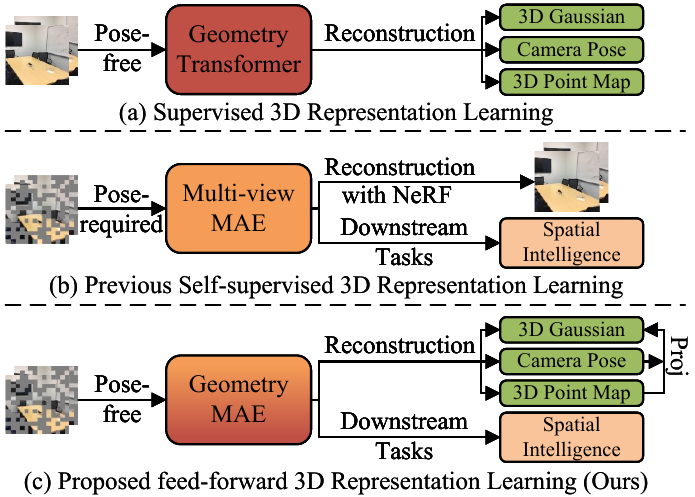}
\captionsetup{font=small,width=1\linewidth}
\caption{
\small{Comparison of three key 3D representation learning paradigms: supervised geometry reconstruction, self-supervised neural reconstruction, and our novel feed-forward framework.
}}
\label{fig:motivation}
\vspace{-15pt}
\end{figure}

Recent advances in visual intelligence highlight the need to retain structured spatial information in learned representations~\cite{cai2025cycle,cai2026unbiased,yang2025changetitans,mao2025prune,xiao2024spatialformer,wang2025visual}. This trend demands representations that unify spatial layouts, semantic contexts, and visual details. Such integration forms the foundation of spatial intelligence, defined as the capacity to construct and reason about the physical world~\cite{yang2025thinking}. This intelligence is essential for embodied agents to navigate, manipulate, and plan within complex environments. Achieving this fundamentally requires robust 3D perception to build comprehensive models of the environment encompassing geometry, appearance, and semantics~\cite{zhu2025spa}. However, deriving such effective 3D representations directly from unposed multi-view images remains an open challenge.

Supervised feed-forward reconstruction methods infer geometry, appearance, and semantics directly from images using ground-truth supervision (Fig.\ref{fig:motivation}(a)). Neural Radiance Fields (NeRFs)\citep{mildenhall2021nerf} and their extensions~\citep{barron2021mip,muller2022instant} achieve high-fidelity novel-view synthesis but require calibrated multi-view images and frequently rely on per-scene optimization. 3D Gaussian Splatting~\citep{kerbl20233d} accelerates training and rendering via explicit primitives, inspiring both pose-aware~\citep{chen2021mvsnerf,xu2024murf,chen2024mvsplat,tang2024lgm} and pose-free pipelines~\citep{jiang2023leap,wang2023pf,smart2024splattr,ye2025pose}. Simultaneously, semantic scene fields~\citep{zhi2021place,peng2021animatable,fan2024large,li2025semanticsplat} advance 3D semantic understanding, whereas geometry-focused approaches predict camera poses or dense point maps to recover the structure of the scene~\citep{wang2024dustr,wang2025vggt,zhang2025flare,leroy2025grounding}. However, most supervised methods require ground-truth geometry or calibration and treat geometry, appearance, and semantics in isolation, lacking the unified representations necessary for spatial intelligence.

Self-supervised approaches mitigate reliance on expensive 3D annotations by constructing geometry priors from unlabeled images. Inspired by 2D representation learning, masked autoencoding~\citep{he2022masked,bao2022beit,dong2025embodiedmae} and contrastive objectives~\citep{chen2020simple,grill2020bootstrap} have been extended to 3D (Fig.\ref{fig:motivation}(b)) to promote cross-view invariance and reconstruction consistency\citep{weinzaepfel2022croco,zhu2023ponderv,zhu2025spa}. While novel-view synthesis frequently serves as a training signal, many methods assume dense video supervision and degrade in sparse-view regimes~\citep{bian2023nope,fu2024colmap}. Recent pose-free self-supervised methods jointly estimate cameras and scene structures~\citep{jiang2025rayzer,kang2025selfsplat}, highlighting the importance of a consistent spatial frame. Nevertheless, most methods still suffer from weak geometry induction, limited appearance details, and inconsistencies between geometry and semantics. This motivates frameworks that explicitly couple geometry, appearance, semantics, and camera estimation to establish a consistent perceptual basis for embodied AI.

To address these challenges, we propose \textbf{\textit{UniSplat}} (Fig.~\ref{fig:motivation}(c)), a feed-forward framework designed to learn unified 3D representations from unposed multi-view images. The proposed framework introduces three key innovations. First, a \textit{dual-masking strategy} enforces geometry-aware feature learning by masking tokens in both the encoder and the decoder. By targeting decoder masks toward geometry-rich regions, this strategy strengthens geometry induction from incomplete visual evidence. Second, a \textit{coarse-to-fine Gaussian splatting strategy} hierarchically refines the radiance field. This progressive enhancement of appearance details facilitates the production of high-fidelity visual representations. Finally, a \textit{pose-conditioned recalibration mechanism} enforces consistency between geometry and semantics by interrelating predictions from the decoder. Unlike conventional multi-task learning, in which each head operates independently, the proposed design utilizes estimated camera poses to reproject 3D point maps and semantic maps onto the 2D image plane. Aligning these reprojected maps with the corresponding RGB and semantic predictions ensures cross-task coherence and resolves mismatches between geometry and semantics. Experiments conducted on diverse benchmarks for 3D scene understanding and embodied AI confirm consistent performance gains, thereby validating the generalization ability of the proposed framework.

Our contributions can be summarized as follows:

\begin{itemize}[leftmargin=*,nosep,itemsep=2pt]
    \item We introduce a dual-masking strategy that applies masking to tokens in both the encoder and the decoder. Biasing the decoder masks toward geometry-rich regions encourages the formation of geometry-aware representations from incomplete visual cues.
 
    \item We propose a coarse-to-fine Gaussian splatting strategy that hierarchically refines the radiance field. This strategy enhances consistency between appearance and semantics, producing high-fidelity visual representations.

    \item We design a pose-conditioned recalibration mechanism that reprojects 3D point maps and semantic maps onto the image plane using estimated camera poses. Aligning these projections with RGB and semantic predictions enforces spatial coherence across tasks.

\end{itemize}

%% file: sec/2_related.tex
% \vspace{-3pt}
\section{Related Work}
\label{sec:related}
% \vspace{-3pt}

\subsection{Supervised 3D Representation Learning}
% \vspace{-3pt}
Supervised feed-forward approaches aim to recover geometry, appearance, and semantics in a single pass using explicit supervision such as target-view rendering signals or external priors. These methods allow fast inference without per-scene optimization~\citep{kerbl20233d,lu2024scaffoldgs,qin2024langsplat,zhou2024feature}. A key distinction lies in their dependence on camera poses. \textbf{Pose-required models} assume known intrinsic and extrinsic characteristics during both training and testing. They often rely on epipolar constraints, cost volumes, or pose-conditioned embeddings, achieving strong photometric quality~\citep{chen2021mvsnerf,xu2024murf,charatan2024pixelsplat,chen2024mvsplat,tang2024lgm,xu2024grm,zhang2024gs,xu2024depthsplat}. However, these methods depend on computationally expensive Structure-from-Motion (SfM) preprocessing and may fail when pose estimation is unreliable.

\textbf{Pose-free models} remove pose inputs at inference but still require posed supervision or fixed targets during training. Within this family, research has diverged by task emphasis. Geometry-focused models predict camera parameters or dense point maps aligned with the scene structure~\citep{wang2024dustr,wang2025vggt,zhang2025flare,leroy2025grounding,jiang2025anysplat,li2025vicasplat}. Appearance-focused methods directly predict per-pixel Gaussians in a canonical space, resolving scale by encoding intrinsics~\citep{jiang2023leap,wang2023pf,smart2024splattr,ye2025pose}. Semantics-focused approaches lift 2D vision-language features into a 3D-consistent field, enabling open-vocabulary, view-consistent segmentation~\citep{fan2024large,li2025semanticsplat,sheng2025spatialsplat}. Parallel efforts such as UniForward~\citep{tian2025uniforward} and Uni3R~\citep{sun2025uni3r} attempt to unify geometry, semantics, and appearance in a single feed-forward pipeline with knowledge distillation. While these strands are beginning to converge, the reliance on posed training combined with unposed inference can lead to residual inconsistencies, motivating the development of self-supervised alternatives.

% \vspace{-3pt}
\subsection{Self-supervised 3D Representation Learning}
% \vspace{-3pt}

\begin{figure*}[!ht]
    \centering
    \setlength{\abovecaptionskip}{2pt}
    \includegraphics[width=1.0\textwidth]{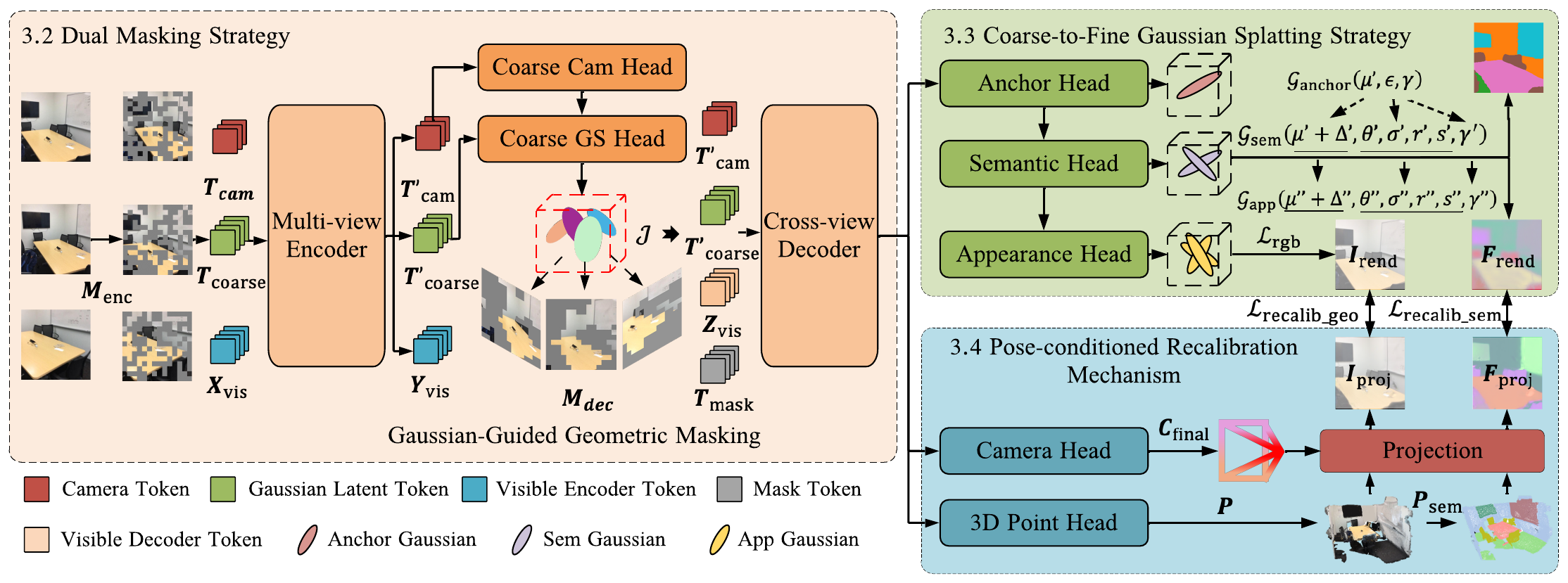}
    \caption{\textbf{Overview of the proposed UniSplat framework.} UniSplat integrates a dual-masking strategy for geometry induction (\S\ref{sec:masking}), a coarse-to-fine Gaussian splatting strategy for appearance refinement (\S\ref{sec:rendering}), and a pose-conditioned recalibration mechanism for geometry–semantic consistency (\S\ref{sec:recalibration}).}
    \label{fig:overview}
    \vspace{-15pt}
\end{figure*}

Self-supervised approaches aim to reduce the reliance on costly 3D labels by learning 3D representations directly from raw multi-view images. 
\textbf{Early methods extended ideas from 2D representation learning}: contrastive learning encouraged view-invariant features~\citep{chen2020simple,grill2020bootstrap}, while masked autoencoding and cross-view completion promoted reconstruction and correspondence~\citep{he2022masked,bao2022beit,weinzaepfel2022croco,zhu2023ponderv,zhu2025spa,dong2025embodiedmae}. These methods improved feature learning, but they often lacked strict global 3D consistency and produced representations that were more view-aligned than spatially grounded.
Another line of work employed \textbf{novel view synthesis} as a self-supervisory signal, where models were trained to render unseen target views and match them photometrically~\citep{bian2023nope,fu2024colmap}. Although this supervision tied predictions more directly to 3D structure and improved geometry–appearance coupling, most approaches assumed known or pre-estimated camera poses or leveraged video metadata to simplify correspondence. Moreover, they typically require dense video streams and re-render nearby frames, which limits robustness in sparse-view settings and constrains applicability to real-world scenarios.

More recently, \textbf{pose-free self-supervised methods} seek to remove this dependency by jointly estimating cameras and scenes directly from raw, unposed image collections. RayZer~\citep{jiang2025rayzer} exemplifies this direction with a transformer-based latent renderer that couples camera and scene recovery in a predict-then-render loop. SelfSplat~\citep{kang2025selfsplat} employs explicit Gaussian splatting, predicting depth and pose with separate modules, which yields interpretable outputs but less coherent alignment. Latent models offer flexibility, while explicit ones provide interpretability and efficient rendering; both highlight a shift toward joint camera–scene learning.
Despite this progress, current pose-free methods still suffer from weak geometry induction, limited appearance detail, and geometry–semantic inconsistencies, underscoring the need for frameworks that explicitly couple all three aspects with camera estimation in a consistent 3D reference frame. 
In this paper, we propose UniSplat, a feed-forward framework that strengthens geometry induction through dual-masking, enhances appearance fidelity via coarse-to-fine Gaussian splatting, enforces geometry–semantic consistency with pose-conditioned recalibration, and combines self-supervised learning with knowledge distillation, yielding robust 3D representations that transfer effectively to scene understanding and embodied AI tasks.

%% file: sec/3_method.tex
\section{Methodology}
\label{sec:methodology}

\subsection{Overview}

We propose UniSplat, a framework for 3D representation learning that takes unposed multi-view images as input. As illustrated in Fig.~\ref{fig:overview}, the model is composed of a transformer encoder and a multi-head decoder. To process the input, a dual-masking strategy (\S\ref{sec:masking}) is applied, which utilizes random masking in the encoder and geometry-aware masking in the decoder to strengthen geometry induction. Subsequently, a coarse-to-fine Gaussian splatting strategy (\S\ref{sec:rendering}) decodes the latent representations produced by the model into multi-level Gaussian parameters. This process renders and reconstructs the RGB appearance and the semantic field of the scene. Finally, a pose-conditioned recalibration mechanism (\S\ref{sec:recalibration}) is introduced to enforce the consistency between 2D and 3D representations. This mechanism leverages the camera parameters predicted by the pose head and the 3D point clouds generated by the 3D point head to project the 3D structures back into the 2D image plane, thereby aligning them with the corresponding 2D predictions. The training objectives are introduced in~\S\ref{sec:objectives}.

\subsection{Dual Masking Strategy}

\label{sec:masking}

To strengthen geometry induction, UniSplat employs a \textbf{dual-masking strategy} inspired by~\citep{wang2023videomae} that applies masking to both encoder and decoder tokens. By targeting decoder masks toward geometry-rich regions, the model is encouraged to infer global 3D structure from incomplete visual cues rather than overfitting to trivial textures.

\noindent\textbf{Stage 1: Initial Masking and Augmented Encoding.}

Given a set of unposed multi-view images $\mathcal{I}\!=\!\{\bm{I}^v\in\mathbb{R}^{H\times W\times 3}\}_{v=1}^V$, we partition each image into patch tokens $\mathcal{X}\!=\!\{\bm{X}^v\in\mathbb{R}^{N_p\times D}\}_{v=1}^V$, where $V$ is the number of views, $N_p$ is the number of patches, and $D$ is the latent dimension. A 2D random mask $\bm{M}_{\text{enc}}^v(\rho_e)$ with masking ratio $\rho_e$ is applied to each $\bm{X}^{v}$:
\begin{equation}
\bm{X}^{v}_\text{vis} = (1 - \bm{M}_{\text{enc}}^v) \odot \bm{X}^{v},
\vspace{-2pt} 
\end{equation}
where $N_{\text{vis}}$ denotes the number of visible patches after masking, typically satisfying $N_{\text{vis}}\!=\!(1-\rho_e)N_p$. This yields the visible token set $\mathcal{X}_\text{vis}\!=\!\{\bm{X}^{v}_\text{vis}\!\in\!\mathbb{R}^{N_{\text{vis}}\times D}\}_{v=1}^V$.
Following~\citep{jiang2025rayzer}, the encoder $\texttt{Enc}$ is augmented with learnable camera tokens $\mathcal{T}_{\text{vis}}\!=\!\{\bm{T}^v_\text{cam}\in\mathbb{R}^{D}\}_{v=1}^V$ and Gaussian latent tokens $\mathcal{T}_{\text{coarse}}\!=\!\{\bm{T}^v_\text{coarse}\in\mathbb{R}^{N_g\times D}\}_{v=1}^V$, where $N_g$ is the number of latent Gaussians. The encoder processes the concatenated token sequence $[\mathcal{X}_\text{vis}, \mathcal{T}_{\text{cam}}, \mathcal{T}_{\text{coarse}}]$ as follows:
\begin{equation}
    [\mathcal{Y}_\text{vis}, \mathcal{T}_{\text{cam}}^{\prime}, \mathcal{T}_{\text{coarse}}^{\prime}] = \texttt{Enc}([\mathcal{X}_\text{vis}, \mathcal{T}_{\text{cam}}, \mathcal{T}_{\text{coarse}}]),
\end{equation}
where $\mathcal{Y}_\text{vis}\!=\!\{\bm{Y}_\text{vis}^v\!\in\!\mathbb{R}^{D \times N_{\text{vis}}}\}_{v=1}^V$ are encoded features, $\mathcal{T}_{\text{cam}}^{\prime}\in\mathbb{R}^{D \times V }$ are updated camera tokens, and $\mathcal{T}_{\text{coarse}}^{\prime}\in\mathbb{R}^{ N_g\times D \times V}$ are updated Gaussian latent tokens.

\noindent\textbf{Stage 2: Gaussian-Guided Geometric Masking.} 
Next, encoded features are used to guide a second geometry-aware masking pass. Updated camera tokens $\mathcal{T}_{\text{cam}}^{\prime}$ are passed to a \textit{Coarse Camera Head}:
\begin{equation}
    \mathcal{C}_\text{coarse} = \texttt{Head}^\text{coarse}_\text{cam}(\mathcal{T}_{\text{cam}}^{\prime}),
\end{equation}
where $\mathcal{C}_\text{coarse} \!=\! \{\bm{C}_{\text{coarse}}^v\!\in\!\mathbb{R}^9\}_{v=1}^V$ encodes the intrinsics and extrinsics of images.
In parallel, Gaussian tokens $\mathcal{T}_{\text{coarse}}^{\prime}$ together with $\mathcal{C}_\text{coarse}$ are passed to a \textit{Coarse Gaussian Head}, forming a preliminary geometric Gaussian field $\mathcal{G}_{\text{geo}}$:
\begin{equation}
    \mathcal{G}_{\text{geo}}(\mu, \sigma, r, s, \beta) = \texttt{Head}^\text{coarse}_\text{gauss}(\mathcal{T}_{\text{coarse}}^{\prime}, \mathcal{C}_\text{coarse}),
\end{equation}
where each Gaussian has the center position $\mu\!\in\! \mathbb{R}^3$, opacity $\sigma\! \in\! \mathbb{R}^{+}$, rotation $r \!\in\! \mathbb{R}^{4}$, scale $s \!\in\! \mathbb{R}^3$, and learnable importance score $\beta \!\in \mathbb{R}^+$.
To identify the most structurally critical regions, we render a geometric importance map $\mathcal{J} \in \mathbb{R}^{ H \times W \times V }$ via alpha blending:
\begin{equation}
    \mathcal{J} = \sum\nolimits_{i=1}^{N_g} \sigma_i \beta_i \prod\nolimits_{j=1}^{i-1}(1-\sigma_j).
\label{eq:importance_map}
\end{equation}
For each patch, an average importance score is obtained by pooling pixel values from $\mathcal{J}$. Patches exceeding the dual masking threshold $\rho_d$ are selected, defining a geometry-aware mask $\bm{M}_{\text{dec}}^v(\rho_d)$. Applying this second mask to the visible tokens from the first stage yields:
\begin{equation}
    \bm{Z}^v_{\text{vis}} = (1 - \bm{M}_{\text{dec}}^v) \odot \bm{Y}^v_{\text{vis}}.
\end{equation}
Subsequently, $\mathcal{Z}_\text{vis}\!=\!\{\bm{Z}^v_{\text{vis}}\}_{v=1}^V$ is fed into the decoder.

By selectively hiding structurally important features, this dual masking forces the decoder to reconstruct them from sparse evidence by reasoning about the underlying 3D spatial structure rather than relying on local texture completion, thereby improving geometry-aware representation learning.
% \vspace{-3pt}
\subsection{Coarse-to-Fine Gaussian Splatting Strategy}
\label{sec:rendering}
% \vspace{-3pt}

A central challenge in Gaussian-based 3D scene understanding is the granularity mismatch between semantic and appearance representations: semantic fields are coarse by nature, while appearance fields require dense, fine-grained primitives to capture textures and lighting.
To reconcile this, UniSplat introduces a \textbf{coarse-to-fine Gaussian splatting strategy} that progressively refines scene representations from global structure to fine detail. 

The decoder takes as input the visible tokens $\mathcal{Z}_\text{vis}$, updated camera tokens $\mathcal{T}_{\text{cam}}'$, Gaussian latent tokens $\mathcal{T}_{\text{coarse}}'$, and learnable masked tokens $\mathcal{T}_{\text{mask}}$. Through a \textbf{multi-head design}, it produces refined Gaussian latent tokens $\mathcal{T}_{\text{coarse}}''$, which are subsequently decoded by hierarchical Gaussian heads for appearance and semantics. 

Building on these outputs, our coarse-to-fine Gaussian splatting proceeds in three stages. 
First, the \textit{Anchor Gaussian Head} predicts anchor Gaussians $\mathcal{G}_{\text{anchor}}(\mu', \epsilon, \gamma )$ from $\mathcal{T}_{\text{coarse}}''$, where $\mu'\in\mathbb{R}^3$ denotes center position, $\epsilon\in\mathbb{R}^{11}$ is the geometric feature, and $\gamma\in\mathbb{R}^{64}$ is the semantic feature. Similar to Scaffold-GS~\citep{lu2024scaffoldgs}, each anchor Gaussian serves as a base from which multiple semantic Gaussians are derived by applying learned position offsets, enabling richer coverage of the local scene context. Next, the \textit{Semantic Gaussian Head} expands each anchor into semantic Gaussians $\mathcal{G}_\text{sem}(\mu'+\Delta', \theta', \sigma', r', s', \gamma')$ by predicting offsets $\Delta'\in\mathbb{R}^3$, coarse appearance attributes $(\theta'\in\mathbb{R}^3, \sigma'\in\mathbb{R}^{+}, r'\in\mathbb{R}^{4}, s'\in\mathbb{R}^{3})$, and semantic features $\gamma'\in\mathbb{R}^{64}$. These semantic Gaussians can be rasterized into 2D maps:
\vspace{-2pt}
\begin{equation}
    \mathcal{S} = \sum\nolimits_{i=1}^{N_s}\sigma'_i \gamma'_i\prod\nolimits_{j=1}^{i-1}(1-\sigma'_j),
\end{equation}
where $N_s$ is the number of semantic Gaussians ($N_s$=10$N_g$ in our implementation). Finally, each semantic Gaussian acts as a new anchor to diffuse a denser set of fine-grained appearance Gaussians $\mathcal{G}_\text{app}$, with refined attributes $(\theta''\!\in\!\mathbb{R}^3, \sigma''\!\in\!\mathbb{R}^{+}, r''\!\in\!\mathbb{R}^{4}, s''\!\in\!\mathbb{R}^{3})$ predicted by the \textit{Appearance Gaussian Head}, predicted offsets $\Delta''\!\in\!\mathbb{R}^3$ and fine-grained semantic features $\gamma''\!\in\!\mathbb{R}^{64}$. The whole process can be summarized as:
\begin{equation}
\begin{aligned}
    \mathcal{G}_{\text{anchor}}(\mu', \epsilon, \gamma ) \Rightarrow \mathcal{G}_\text{sem}(\mu'+\Delta', \theta', \sigma', r', s', \gamma') \\ 
    \Rightarrow \mathcal{G}_\text{app}(\mu''+\Delta'', \theta'', \sigma'', r'', s'', \gamma'').
\end{aligned}
\end{equation}
% \vspace{-2pt}
By progressively refining anchor, semantic, and appearance Gaussians, this strategy resolves the granularity mismatch between coarse semantics and fine textures.

% \vspace{-3pt}
\subsection{Pose-conditioned Recalibration Mechanism}
\label{sec:recalibration}

To ensure geometric–semantic consistency, UniSplat introduces a \textbf{pose-conditioned recalibration mechanism}. This component aligns predictions from different heads by reprojecting 3D outputs into the image plane using estimated camera parameters and minimizing their discrepancy with 2D renderings, thereby enforcing cross-task coherence. 

For pose-conditioned recalibration, the decoder outputs updated visible tokens $\mathcal{Z}_{\text{vis}}''$, camera tokens $\mathcal{T}_{\text{vis}}''$, and masked tokens $\mathcal{T}_{\text{mask}}$, which are further consumed by a \textit{Point Head} and a \textit{Camera Head}. The \textit{Point Head} uses a Dense Prediction Transformer (DPT) to regress per-view 3D point maps $\mathcal{P}=\{\bm{P}^v\!\in\! \mathbb{R}^{H \!\times\! W \!\times\! 3}\}_{v=1}^{V}$, while the \textit{Camera Head} predicts refined camera parameters $\mathcal{C}_{\text{final}}\!=\! \{\bm{C}_{\text{final}}^v\!\in\!\mathbb{R}^9\}_{v=1}^V$. These predictions provide the geometric cues required by the recalibration mechanism.

\begin{figure}[!t]
% \vspace{-2pt}
\centering
\setlength{\abovecaptionskip}{2pt}
\includegraphics[width=1\linewidth]{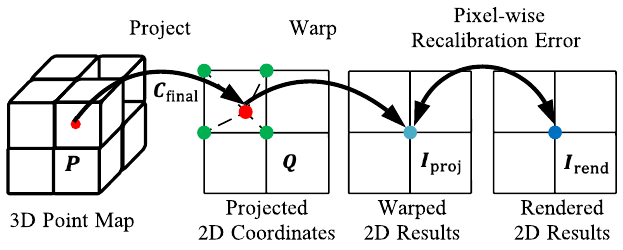}
\captionsetup{font=small,width=1\linewidth}
\caption{
\small{Pose-conditioned recalibration mechanism (\S\ref{sec:recalibration}).
}}
\label{fig:projection}
\vspace{-15pt}
\end{figure}

UniSplat produces two complementary types of 3D predictions: 3D Gaussian fields ($\mathcal{G}_{\text{app}}$ for appearance and $\mathcal{G}_\text{sem}$ for semantics) and 3D point maps $\mathcal{P}$. To ensure consistency among these predictions, the recalibration mechanism projects 3D maps back into 2D, as shown in Fig.~\ref{fig:projection}. Since the projected coordinates $\mathcal{Q}\!=\!\{\bm{Q}^v\!\in\!\mathbb{R}^{H\times W \times 2}\}_{v=1}^{V}$ are continuous, values are interpolated via differentiable bilinear sampling from four neighboring pixels to obtain warped results $\mathcal{I}_{\text{proj}}\!=\!\{\bm{I}_{\text{proj}}^v\!\in\!\mathbb{R}^{H\times W\times 3}\}_{v=1}^{V}$, which are then compared with Gaussian-rendered results $\mathcal{I}_{\text{rend}}\!=\!\{\bm{I}_{\text{rend}}^v\!\in\!\mathbb{R}^{H\times W\times 3}\}_{v=1}^{V}$ to compute the pixel-wise recalibration error, as shown in Fig.~\ref{fig:projection}. This procedure forces appearance, semantics, and geometry to converge toward a consistent scene representation.

\noindent\textbf{Geometric Recalibration.} 
For each view $v$, the 3D point map $\bm{P}^{v}$ is projected to 2D using the refined camera parameters $\bm{C}_\text{final}$. We then measure consistency with the rendered RGB image from the appearance field $\mathcal{G}\text{app}$ via a pose-conditioned reprojection loss:
\vspace{-2pt} 
\begin{equation}
    \mathcal{L}_{\text{recalib\_geo}} = \sum\limits_{v=1}^{V}\sum\limits_{j=1}^{H\times W}\|\bm{I}_{\text{rend}, j}^{v}-\texttt{Proj}(\bm{C}_{\text{final}}^{v},\bm{P}_j^{v})\|,
\vspace{-2pt} 
\end{equation}
where $\texttt{Proj}(\cdot)$ is the projection operator.

\noindent\textbf{Semantic Recalibration.} 
To align semantic information, we build a 3D semantic point map $\mathcal{P}_{\text{sem}}$ by associating each 3D point in $\mathcal{P}_{\text{sem}}$ with its semantics. Similarly, projecting $\mathcal{P}_{\text{sem}}$ to 2D yields a semantic projection $\mathcal{F}_{\text{proj}}\!=\! \{\bm{F}_{\text{proj}}^v\!\in\!\mathbb{R}^{H\times W\times 64}\}_{v=1}^{V}$, which is aligned with the rendered semantic map $\mathcal{F}_{\text{rend}}\!=\!\{\bm{F}_{\text{rend}}^v \}_{v=1}^{V} $ from $\mathcal{G}_\text{sem}$:
\begin{equation}
    \mathcal{L}_{\text{recalib\_sem}} = \sum\limits_{v=1}^{V}\sum\limits_{j=1}^{H\times W} \big( 1 - \frac{\bm{F}^{v}_{\text{proj, j}} \cdot \bm{F}^{v}_{\text{rend, j}}}{\|\bm{F}^{v}_{\text{proj, j}}\| \cdot \|\bm{F}^{v}_{\text{rend, j}}\|} \big).
\end{equation}

The overall recalibration objective combines both terms:
\begin{equation}
    \mathcal{L}_{\text{recalib}} = \mathcal{L}_{\text{recalib\_geo}} + \mathcal{L}_{\text{recalib\_sem}}.
\end{equation}
By enforcing both geometric and semantic alignment in the 2D image plane, the recalibration mechanism ensures consistency between geometry and semantics, which is crucial for producing coherent 3D scene representations.

% \vspace{-3pt}
\subsection{Training Objectives}
\label{sec:objectives}

Self-supervision from input views provides an essential learning signal, but it alone is insufficient for reliable 3D modeling. Thus, we adopt a composite objective that combines self-supervision with knowledge distillation, leveraging geometric and semantic priors from large-scale pre-trained foundation models to strengthen learning while avoiding the need for expensive 3D labels.
The overall objective is a weighted sum of four terms: 
\begin{equation}
    \mathcal{L}_{\text{total}} = \mathcal{L}_{\text{rgb}} + \mathcal{L}_{\text{sem}} + \mathcal{L}_{\text{geo}} + \mathcal{L}_{\text{recalib}}.
\end{equation}

\noindent\textbf{Photometric Reconstruction Loss} ensures that the rendered appearance $\bm{I}_{\text{rend}}^v$ from $\mathcal{G}_\text{app}$ matches the input views $\bm{I}^v$. It combines an L1 loss with the LPIPS perceptual metric for image quality~\cite{zhang2018unreasonable}:
\vspace{-3pt}
\begin{equation}
    \mathcal{L}_{\text{rgb}} = \sum\limits_{v=1}^{V} \big( \| \bm{I}_{\text{rend}}^v - \bm{I}^v \| + \mathcal{L}_{\text{LPIPS}}(\bm{I}_{\text{rend}}^v, \bm{I}^v) \big).
\vspace{-3pt}
\end{equation}

\noindent\textbf{Semantic Distillation Loss} transfers open-vocabulary semantic knowledge from a frozen 2D vision–language model (VLM) into the 3D semantic Gaussians. For each view $v$, we extract a semantic feature map $F^v_{\text{VLM}}$ from the input image $\bm{I}^v$ using the VLM’s image encoder (\textit{e.g.}, LSeg~\citep{li2022languagedriven}) and align it with the rendered semantic feature map $F^v_{\text{render}}$. The loss is defined as one minus the cosine similarity between these 2D semantic features:
\begin{equation}
    \mathcal{L}_{\text{sem}} = \sum\limits_{v=1}^{V}\sum\limits_{j=1}^{H\times W} \big(1 - \frac{\bm{F}^{v}_{\text{render, j}} \cdot \bm{F}^{v}_{\text{vlm, j}}}{\|\bm{F}^{v}_{\text{render, j}}\| \cdot \|\bm{F}^{v}_{\text{vlm, j}}\|} \big).
\end{equation}

\noindent\textbf{Geometric Prior Loss.} Following~\citep{jiang2025anysplat}, we transfer geometric knowledge from a frozen VGGT~\citep{wang2025vggt} teacher, which provides pseudo ground-truth camera parameters $\tilde{\bm{C}}^{v}$ and point maps $\tilde{\bm{P}}^{v}$, to regularize camera estimation and strengthen 3D structure learning. Camera parameters are regularized via a Huber loss:
\vspace{-3pt}
\begin{equation}
    \mathcal{L}_\text{pose}=\sum\nolimits_{v=1}^{V} \big\|\tilde{\bm{C}}^{v}-\bm{C}_{\text{final}}^{v}\big\|_\epsilon.
\vspace{-3pt}
\end{equation}
3D scene geometry is distilled using:
\begin{equation}
\begin{aligned}
    \mathcal{L}_\text{point}\!=\!\sum\limits_{v=1}^{V} \sum\limits_{j=1}^{H\times W} \!\widetilde{\bm{U}}^{v}_j\cdot\|\tilde{\bm{P}}^{v}_j-\bm{P}^{v}_j\|  
     +\|\widetilde{\bm{U}}^{v}_j-\bm{U}^{v}_j\|,
\vspace{-5pt} 
\end{aligned}
\end{equation}
where $\widetilde{\bm{U}}^{v}_j$ denotes the confidence score of the point map.
The full geometric prior loss combines these terms:
\begin{equation}
    \mathcal{L}_{\text{geo}} = \lambda_{\text{pose}} \mathcal{L}_\text{pose} + \lambda_{\text{point}} \mathcal{L}_\text{point},
\vspace{-2pt} 
\end{equation}
where $\lambda_{\text{pose}}$ and $\lambda_{\text{point}}$ are set to 10.0 and 1.0.

%% file: sec/4_experiments.tex
\section{Experiments}

\begin{table*}[!ht]%[!h]
    
    \centering
    \setlength{\abovecaptionskip}{2pt}
    \caption{\textbf{Quantitative comparison on ScanNet~\cite{dai2017scannet}.} (\S\ref{sec:3d_results}) We evaluate performance on novel view synthesis (NVS), depth estimation (DE), and open-vocabulary semantic segmentation (OVSS).}
    
    \resizebox{0.99\textwidth}{!}{
    \setlength\tabcolsep{3pt}
    \renewcommand\arraystretch{1.0}
    \begin{tabular}{l|| cc | cc | cc | cc | ccc}
    \hline \thickhline \rowcolor{mygray}
     &  & & \multicolumn{4}{c|}{Source View} & \multicolumn{5}{c}{Target View} \\
    \cline{2-12}
    \rowcolor{mygray}  &\multicolumn{2}{c|}{Recon. Time$\downarrow$} & \multicolumn{2}{c|}{OVSS} & \multicolumn{2}{c|}{DE} & \multicolumn{2}{c|}{OVSS} & \multicolumn{3}{c|}{NVS} \\
    \cline{4-12}
    \rowcolor{mygray} \multirow{-3}{*}{Methods}& SfM& Per-Scene & mIoU$\uparrow$ & Acc.$\uparrow$ & rel$\downarrow$ & $\tau\uparrow$ & mIoU$\uparrow$ & Acc.$\uparrow$ & PSNR$\uparrow$ & SSIM$\uparrow$ & LPIPS$\downarrow$ \\
    \hline\hline
    LSeg~\cite{li2022languagedriven} & N/A & N/A & 0.4701 & 0.7891 & - & - & 0.4819 & 0.7927 & - & - & - \\
    NeRF-DFF~\cite{kobayashi2022decomposing} & 20.52s & 1min & 0.4540 & 0.7173 & 27.68 & 9.61 & 0.4037 & 0.6755 & 19.86 & 0.6650 & 0.3629 \\
    Feature-3DGS~\cite{zhou2024feature} & 20.52s & 18mins & 0.4453 & 0.7276 & 12.95 & 21.07 & 0.4223 & 0.7174 & 24.49 & 0.8132 & 0.2293 \\
    \hline
    pixelSplat~\cite{charatan2024pixelsplat} & 20.52s & 0.064s & - & - & - & - & - & - & 24.89 & 0.8392 & 0.1641 \\
    LSM~\cite{fan2024large} & \multicolumn{2}{c|}{0.108s} & 0.5034 & 0.7740 & 3.38 & 67.77 & 0.5078 & 0.7686 & 24.39 & 0.8072 & 0.2506 \\\hline
    Ours & \multicolumn{2}{c|}{0.041s} & \textbf{0.5563} & \textbf{0.8277} & \textbf{3.10} & \textbf{69.13} & \textbf{0.5625} & \textbf{0.8334} & \textbf{25.65} & \textbf{0.8782} & \textbf{0.1353} \\
    \hline
    \end{tabular}
    }
    
    \label{tab:semantic_reconstruction}
    \vspace{-10pt}
\end{table*}

\subsection{Experimental Setup}
To comprehensively evaluate the effectiveness of our unified 3D representation, we test UniSplat in two distinct domains: (1) \textbf{traditional 3D vision tasks} to assess the quality of scene reconstruction, and (2) \textbf{embodied AI tasks} to evaluate the utility of the learned features as a visual backbone for downstream robotic control policies.

\noindent\textbf{3D Vision Tasks.} Following LSM, we evaluate UniSplat's scene understanding capabilities on 40 unseen scenes from the ScanNet~\citep{dai2017scannet} dataset. We assess five core tasks: Novel View Synthesis (PSNR, SSIM, LPIPS), Open-Vocabulary 3D Segmentation (mIoU, mAcc), Depth Estimation (Abs Rel, Inlier Ratio), relative pose estimation (AUC), and cross-dataset generalization.

\noindent\textbf{Embodied AI Tasks.} We use the pre-trained ViT encoder from UniSplat as a frozen feature extractor and evaluate it on the largest-scale embodied intelligence benchmark, which spans 268 tasks across 8 simulators.
The evaluation covers both \textbf{single-task} (VC-1~\citep{majumdar2023where}, Franka Kitchen~\citep{gupta2019relay}, Meta-World~\citep{yu2020meta}) and \textbf{language-conditioned multi-task} (RLBench~\citep{james2020rlbench}, LIBERO~\citep{liu2023libero}) scenarios, utilizing a variety of policies including MLPs, Diffusion, and Transformers.
We compare against a diverse set of leading visual representation learning models, including \textbf{vision-centric} (MAE~\citep{he2022masked}, DINOv2~\citep{oquab2023dinov2}), \textbf{multi-modal} (CLIP~\citep{radford2021learning}, EVA~\citep{fang2023eva}, InternViT~\citep{chen2024internvl}), and \textbf{embodied-specific} (VC-1~\citep{majumdar2023where}, MVP~\citep{radosavovic2023real}, SPA~\citep{zhu2025spa}) approaches.
Please refer to the Appendix for details.

\noindent\textbf{Implementation Details.}
UniSplat is built upon a ViT-L backbone pre-trained on ScanNet and ScanNet++~\citep{yeshwanth2023scannet++}, equipped with a multi-task decoder. To circumvent explicit 3D supervision, we leverage LSeg and VGGT teachers to generate pseudo ground-truth semantics and geometry. All experiments are optimized using AdamW with a base learning rate of $1\times10^{-4}$ and a 30-epoch warm-up schedule. Training is conducted for 300 epochs on 8 NVIDIA A100 GPUs. For fair comparison with baseline methods, the input resolution is fixed at $256\times256$. The dual-masking strategy adopts masking ratios of 0.5 for both encoder and decoder. Each anchor Gaussian yields 10 derived Gaussians, and the number of coarse Gaussian tokens $T_\text{coarse}$ is set to 256. %The loss weights are configu as follows: $\lambda_\text{rgb}\!=\!1.0$, $\lambda_\text{LPIPS}\!=\!0.05$, $\lambda_\text{sem}\!=\!0.3$, $\lambda_\text{geo}\!=\!1.2$, $\lambda_{pose}\!=\!10.0$, $\lambda_\text{point}\!=\!1.5$, $\lambda_\text{recalib\_geom}\!=\!0.001$, $\lambda_\text{recalib\_geo}\!=\!0.5$, and $\lambda_\text{recalib}\!=\!1.0$.

% \vspace{-3pt}
\subsection{Results on 3D Vision Tasks}
\label{sec:3d_results}
% \vspace{-3pt}

Table~\ref{tab:semantic_reconstruction} shows results compared to pose-based baselines (NeRF-DFF~\citep{kobayashi2022decomposing}, Feature-3DGS~\citep{zhou2024feature}, pixelSplat~\citep{charatan2024pixelsplat}) and the pose-free baseline LSM~\citep{fan2024large}.

\noindent\textbf{Open-Vocabulary 3D Segmentation.} As presented in Table~\ref{tab:semantic_reconstruction} and Figure~\ref{fig:qualitative_nvs_scannet}, UniSplat sets a new pose-free state-of-the-art. On source views, it reaches 0.5563 mIoU and 0.8277 mAcc, surpassing LSM by +5.3/+5.4 points. On target views, it attains 0.5625 mIoU and 0.8334 mAcc, improving over LSM by +5.5/+6.5 points and exceeding the 2D LSeg baseline while providing cross-view consistency that 2D methods lack.

\noindent\textbf{Novel View Synthesis.} Without SfM or per-scene fitting, UniSplat delivers the best image quality among compa methods: 25.65 PSNR, 0.8782 SSIM, and 0.1353 LPIPS on target views. This outperforms LSM, the generalizable pixelSplat that assumes known cameras, and the optimized Feature-3DGS. Gains indicate that geometry-aware masking and progressive semantic-to-appearance rendering yield sharper, more faithful novel views. Furthermore, to evaluate rendering quality, we train our model on the RealEstate10K~\citep{zhou2018stereo} datasets. Figure~\ref{fig:qualitative_nvs_re10k} shows a qualitative comparison of the novel view synthesis on RealEstate10K. %More details and results are provided in the Appendix.

\noindent\textbf{Depth Estimation.} UniSplat achieves best performance on source views, improving over LSM. Methods optimized per scene remain far behind in geometry. We provide visualizations of feature maps and depth maps in Fig.~\ref{fig:feature_depth_vis_scannet}.

\begin{table*}[!t]%[!h]
\centering
\setlength{\abovecaptionskip}{2pt}
% \vspace{2pt}
\caption{\textbf{Quantitative comparison on various embodied AI tasks.} (\S\ref{sec:embodied_results}) The number in parentheses denotes the number of tasks.}
\resizebox{0.99\linewidth}{!}{
\setlength\tabcolsep{3pt}
\begin{tabular}{l|l|cc|ccc|cccc}\hline \thickhline \rowcolor{mygray}
\rowcolor{mygray} \multicolumn{2}{r|}{Methods} &\multicolumn{2}{c|}{\textit{Vision-Centric}} &\multicolumn{3}{c|}{\textit{Multi-Modal}} &\multicolumn{4}{c}{\textit{Embodied-Specific}} \\
\rowcolor{mygray} \multicolumn{2}{l|}{Benchmarks}  &MAE &DINOv2 &CLIP &EVA & InternViT &MVP &VC-1 &SPA & Ours  \\\hline\hline
\multirow{4}{*}{\rotatebox{0}{VC-1}} &AD (2)  &$58.0_{\pm2.0}$ &$47.3_{\pm3.1}$  &$48.7_{\pm3.1}$ &$58.0_{\pm6.0}$ &$53.3_{\pm3.1}$ &$53.3_{\pm4.2}$ &$54.0_{\pm4.0}$ &$60.0_{\pm4.0}$ & $\mathbf{{61.7_{\pm4.3}}}$ \\
&MW (5)  &$90.0_{\pm4.6}$ &$84.0_{\pm3.7}$ &$77.1_{\pm3.2}$ &$90.7_{\pm0.9}$ &$84.0_{\pm3.7}$ &$93.6_{\pm5.2}$ &$87.5_{\pm3.8}$ &$93.3_{\pm2.0}$ & $\mathbf{94.3_{\pm3.1}}$ \\
&DMC (5)  &$74.4_{\pm1.8}$ &$64.5_{\pm2.5}$ &$53.9_{\pm3.6}$ &$62.7_{\pm2.8}$ &$53.3_{\pm0.4}$ &$69.4_{\pm2.6}$ &$65.3_{\pm3.6}$ &$71.1_{\pm5.0}$ & $\mathbf{75.8_{\pm4.5}}$ \\
&TF (2)  &$73.0_{\pm0.5}$ &$68.5_{\pm0.4}$ &$56.1_{\pm1.6}$ &$67.2_{\pm0.2}$ &$65.2_{\pm1.6}$ &$73.2_{\pm0.8}$ &$70.9_{\pm1.1}$ &$73.6_{\pm2.0}$ & $\mathbf{75.6_{\pm1.7}}$ \\ \hline
\multirow{2}{*}{\rotatebox{0}{RLBench}} &Group 1 (35)  &78.3 &78.2 &76.8 &75.2 &74.1 &76.2 &80.1 &80.5 & \textbf{81.2} \\
&Group 2 (36)  &57.7 &56.1 &55.7 &57.0 &54.9 &56.3 &55.7 &61.2 & \textbf{63.3} \\\hline
\multicolumn{2}{c|}{Meta-World (48)} &$67.8_{\pm1.7}$ &$56.3_{\pm0.6}$ &$66.7_{\pm1.7}$ &$63.7_{\pm1.3}$ &$57.5_{\pm1.7}$ &$66.4_{\pm1.7}$ &$68.6_{\pm1.5}$ &$69.2_{\pm1.7}$ & $\mathbf{70.9_{\pm1.3}}$ \\\hline
\multirow{5}{*}{\rotatebox{0}{LIBERO}} &Object (10)  &$71.7_{\pm13.1}$ &$64.7_{\pm9.9}$ &$50.2_{\pm7.0}$ &$73.2_{\pm6.0}$ &$67.7_{\pm6.0}$ &$63.7_{\pm4.8}$ &$69.7_{\pm7.2}$ &$76.7_{\pm5.3}$ &$\mathbf{78.4_{\pm6.1}}$ \\
&Spatial (10) &$57.2_{\pm2.9}$ &$36.3_{\pm11.8}$ &$32.2_{\pm0.6}$ &$59.3_{\pm7.7}$ &$48.3_{\pm6.4}$ &$58.0_{\pm6.2}$ &$50.5_{\pm7.5}$ &$50.0_{\pm3.8}$ &$\mathbf{59.7_{\pm5.8}}$ \\
&Goal (10)  &$54.3_{\pm6.0}$ &$22.2_{\pm2.3}$ &$30.3_{\pm3.2}$ &$56.8_{\pm2.9}$ &$58.8_{\pm4.5}$ &63.8$_{\pm2.8}$ &57.5$_{\pm6.6}$ &65.3$_{\pm2.5}$ &$\mathbf{67.3_{\pm2.3}}$ \\
&10 (10) &$41.2_{\pm4.5}$ &$28.3_{\pm3.0}$ &$27.5_{\pm3.9}$ &$43.3_{\pm2.8}$ &$38.2_{\pm1.3}$ &$39.0_{\pm0.9}$ &$39.7_{\pm3.5}$ &$40.2_{\pm3.6}$ &$\mathbf{42.4_{\pm3.5}}$ \\
&90 (90) &$29.9_{\pm2.0}$ &$27.5_{\pm2.2}$ &$29.4_{\pm2.0}$ &$31.3_{\pm2.3}$ &$23.8_{\pm1.8}$ &$32.1_{\pm3.5}$ &$30.6_{\pm3.3}$ &$32.2_{\pm1.6}$ &$\mathbf{34.7_{\pm2.7}}$ \\\hline
\multicolumn{2}{c|}{Franka-Kitchen (5)} &$42.7_{\pm2.6}$ &$40.9_{\pm6.4}$ &$30.8_{\pm3.3}$ &$37.3_{\pm1.3}$ &$28.5_{\pm1.7}$ &$34.3_{\pm6.1}$ &$37.5_{\pm3.5}$ &$40.6_{\pm1.9}$ &$\mathbf{44.5_{\pm2.6}}$ \\\hline

\end{tabular}
}
\label{tab:embodied-tasks}
\vspace{-5pt}
\end{table*}

\begin{table}[!t]
\footnotesize
\centering
\setlength{\abovecaptionskip}{2pt}

\caption{\textbf{Quantitative comparison of relative pose estimation on RealEstate10K~\cite{zhou2018stereo} and ACID~\cite{liu2021infinite}.} (\S\ref{sec:3d_results})}
\resizebox{1.0\linewidth}{!}{
\begin{tabular}{l|ccc|ccc}
    \hline\thickhline
    \rowcolor{mygray}
     & \multicolumn{3}{c}{\textbf{RealEstate10K}} & \multicolumn{3}{c}{\textbf{ACID}}  \\
    \cline{2-4} \cline{5-7}  
    
    \rowcolor{mygray} \textbf{Methods} &  5$^\circ$ $\uparrow$ & 10$^\circ$ $\uparrow$ & 20$^\circ$ $\uparrow$ &  5$^\circ$ $\uparrow$ &  10$^\circ$ $\uparrow$ &  20$^\circ$ $\uparrow$  \\
    \hline \hline
    DUSt3R & 0.329 & 0.537 & 0.691 & 0.113 & 0.273 & 0.469 \\ 
    MASt3R & 0.351	& 0.557 & 0.701	& 0.159 & 0.362 & 0.524 \\
    NoPoSplat & 0.568 & 0.737 & 0.839 & 0.342 & 0.504 & 0.653 \\
    SelfSplat  & 0.223 & 0.413 & 0.589 & 0.213 & 0.372 & 0.541 \\\hline
    \textbf{Ours} & \textbf{0.607} & \textbf{0.748}	& \textbf{0.842} & \textbf{0.354} & \textbf{0.516}	& \textbf{0.661} \\
    \hline
    \end{tabular}
}
    \label{tab:pose_estimation}
    \vspace{-5pt}
    
\end{table}

\begin{table}[!t]
\footnotesize
\centering
\setlength{\abovecaptionskip}{2pt}
\setlength{\tabcolsep}{4pt}
\caption{\textbf{Quantitative comparison of Cross-dataset Generalization on ACID~\cite{liu2021infinite} and DTU~\cite{jensen2014large}.} (\S\ref{sec:3d_results}) All models are trained on RealEstate10K~\citep{zhou2018stereo} (indoor scenes) and evaluated on ACID (outdoor scenes) and DTU (object-centric scenes) datasets.}
\resizebox{1.0\linewidth}{!}{
\begin{tabular}{l|ccc|ccc}
    \hline\thickhline
    \rowcolor{mygray}
     & \multicolumn{3}{c}{\textbf{ACID}} & \multicolumn{3}{c}{\textbf{DTU}}  \\
    \cline{2-4} \cline{5-7}  
    
    \rowcolor{mygray} \textbf{Methods} &  PSNR$\uparrow$ & SSIM$\uparrow$ & LPIPS$\downarrow$ &  PSNR$\uparrow$ & SSIM$\uparrow$ & LPIPS$\downarrow$  \\
    \hline \hline
    pixelSplat & 25.477 & 0.770 & 0.207 & 15.067 & 0.539 & 0.341 \\ 
    MVSplat & 25.525	& 0.773 & 0.199	& 14.542 & 0.537 & 0.324 \\
    NoPoSplat & 25.765 & 0.776 & 0.199 & \textbf{17.899} & \textbf{0.629} & \textbf{0.279} \\
    SelfSplat  & 22.204 & 0.686 & 0.316 & 13.249 & 0.434 & 0.441 \\\hline
    \textbf{Ours} & \textbf{25.983} & \textbf{0.786}	& \textbf{0.188} & \underline{16.852} & \underline{0.587}	& \underline{0.269} \\
    \hline
    \end{tabular}
}
    \label{tab:cross_dataset}
    \vspace{-5pt}
    
\end{table}

\noindent\textbf{Relative Pose Estimation.} We evaluate the relative pose estimation performance on the RealEstate10K and ACID datasets, following \citep{ye2025pose}. As shown in Table~\ref{tab:pose_estimation}, our method outperforms all baselines across all thresholds, demonstrating the effectiveness of our 3D representation in capturing accurate camera poses from unposed multi-view images.

\noindent\textbf{Cross-Dataset Generalization.} Trained on RealEstate10K and evaluated on ACID and DTU, UniSplat generalizes well across domains (Table~\ref{tab:cross_dataset}), achieving top PSNR / SSIM / LPIPS without extra tuning. This demonstrates robust geometry and appearance transfer under varied conditions.

\subsection{Results on Embodied AI Tasks}
\label{sec:embodied_results}

As shown in Table~\ref{tab:embodied-tasks}, UniSplat consistently outperforms vision-centric (MAE, DINOv2), multi-modal (CLIP, EVA, InternViT), and embodied-specific (MVP, VC-1, SPA) baselines. On VC-1, UniSplat attains top scores across AD, MW, DMC, and TF. It sets new records on RLBench (81.2\%/63.3\% for Group~1/2) and Meta-World (70.9\%), while achieving strong gains in all LIBERO splits, including 78.4\% on Object, 59.7\% on Spatial, and 67.3\% on Goal, with robust results on LIBERO-90. On Franka Kitchen, it reaches 44.5\%, surpassing prior methods. These results show that the unified 3D representation transfers effectively to varied control tasks without task-specific tuning.

\begin{figure}[!ht]%[!h]
\setlength{\abovecaptionskip}{2pt}
\centering
\includegraphics[width=0.475\textwidth]{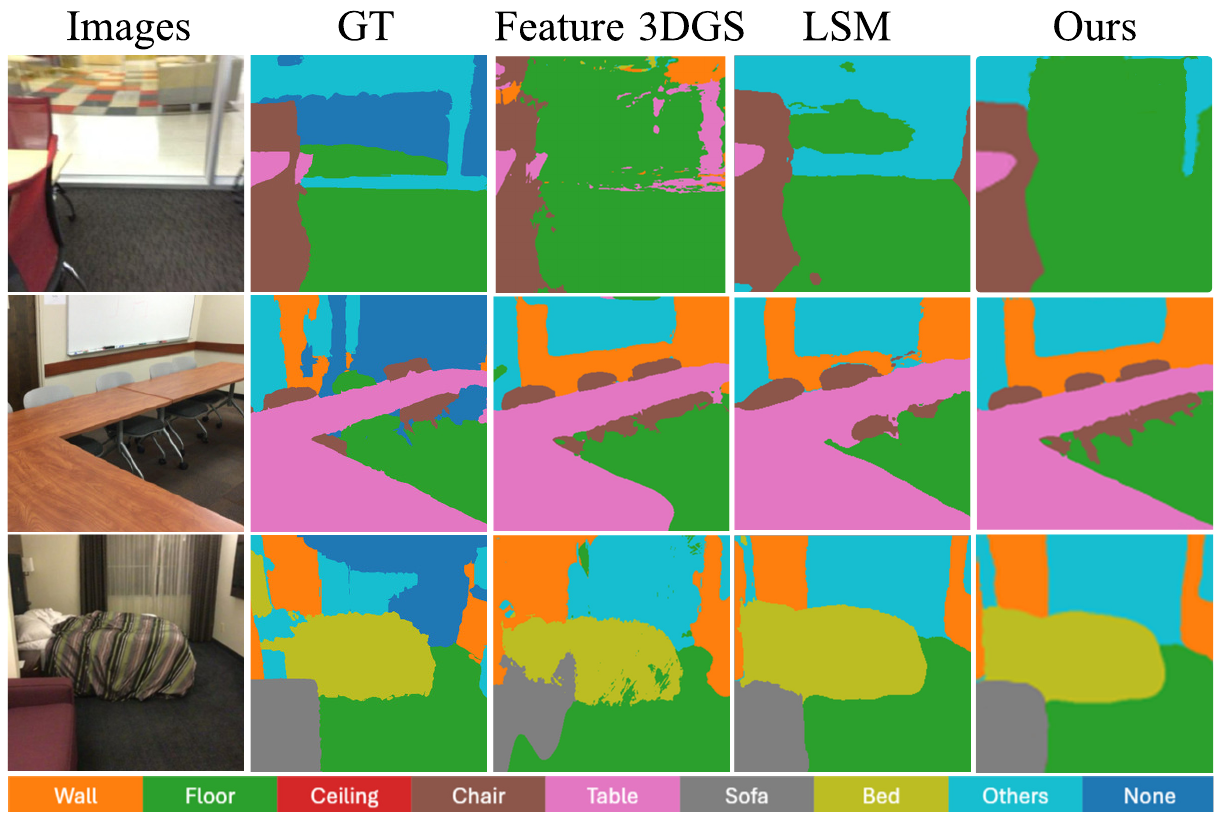}
\caption{\textbf{Qualitative comparison of novel-view segmentation on ScanNet.} (\S\ref{sec:3d_results}) Follwing ~\citep{fan2024large}, we map thousands of ScanNet labels into 8 common categories for visualization. The dense 2D GT labels are obtained by projecting sparse 3D annotations, so some regions are inevitably incomplete or missing.}
\label{fig:qualitative_nvs_scannet}
\vspace{-5pt}
\end{figure}

\begin{figure}[!ht]%[!h]

\setlength{\abovecaptionskip}{2pt}
\centering
\includegraphics[width=0.47\textwidth]{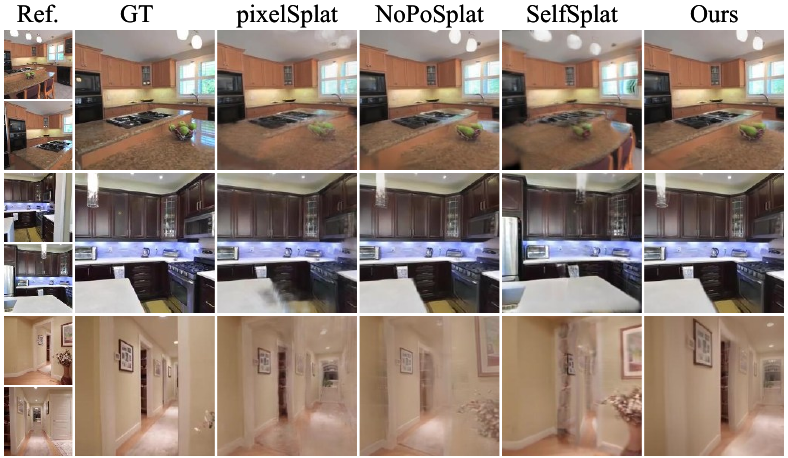}
\caption{\textbf{Qualitative comparison of NVS on RE10K.} (\S\ref{sec:3d_results})}
\label{fig:qualitative_nvs_re10k}
\vspace{-3pt}
\end{figure}

\begin{figure}[!ht]%[!h]
\setlength{\abovecaptionskip}{2pt}
\centering
\includegraphics[width=0.45\textwidth]{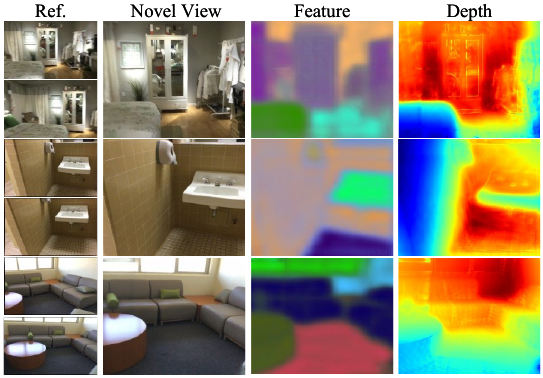}
\caption{\textbf{Visualizations of feature and depth on ScanNet.}}
\label{fig:feature_depth_vis_scannet}
\vspace{-3pt}
\end{figure}

\subsection{Ablation Studies}
\label{sec:ablation}

\begin{table}[!ht]%[!ht]

\centering
\setlength{\abovecaptionskip}{2pt}
\renewcommand{\arraystretch}{1.0}
\caption{\textbf{Ablation on core components.} (\S\ref{sec:ablation}) Each row removes one component to evaluate its contribution.}

\resizebox{\linewidth}{!}{
\begin{tabular}{l|cc|c|c}
\hline
\thickhline
\rowcolor{mygray} 
Variant & mIoU$\uparrow$ & Acc.$\uparrow$ & PSNR$\uparrow$ & rel$\downarrow$ \\
\hline\hline
Full UniSplat & \textbf{0.5625} & \textbf{0.8334} & \textbf{25.65} & \textbf{3.10} \\
w/o Self-sup. & 0.5263 & 0.8110 & 24.40 & 3.74 \\
w/o Dual Mask & 0.5462 & 0.8275 & 24.74 & 3.27 \\ 
w/o Coarse-to-Fine & 0.5374 & 0.8239 & 24.93 & 3.34 \\
w/o $\mathcal{L}_{\text{recalib}}$ & 0.5287 & 0.8186 & 24.35 &  3.52 \\
w/o $\mathcal{L}_{\text{sem}}$ & 0.0214 & 0.0811 & 24.82 &  3.24 \\
w/o $\mathcal{L}_{\text{geo}}$ & 0.5442 & 0.8261 & 24.23 &  3.59 \\
\hline
\end{tabular}
}
\label{tab:ablation_core}
\vspace{-10pt}
\end{table}

\noindent\textbf{Ablation on Key Components.} Table~\ref{tab:ablation_core} shows that every component matters. Removing semantic distillation collapses segmentation while keeping appearance nearly intact, underscoring its role for open-vocabulary semantics. Camera recalibration and geometric prior losses are critical for geometry and rendering. Geometry-aware dual mask and coarse-to-fine Gaussians splatting strategy yield consistent gains. Disabling self-supervised learning for appearance degrades all metrics.

\noindent\textbf{Scale Up with More Training Data.} Table~\ref{tab:ablation_data_scale} demonstrates that expanding the training dataset can significantly enhance model generalization and robustness. Larger, diverse datasets expose the model to varied patterns, reducing overfitting and improving performance on unseen data.

\noindent\textbf{Ablation on the Number of Input Views.} As shown in Table~\ref{tab:ablation_view_num}, increasing input views improves both segmentation and reconstruction. More views provide richer geometric and appearance cues, enhancing scene understanding and rendering quality. Gains diminish beyond 8 views, indicating that while additional viewpoints help, the marginal benefit reduces once sufficient coverage is achieved.

\begin{table}[!t]
\centering
\setlength{\abovecaptionskip}{2pt}
\caption{\textbf{Ablation on data scale.} (\S\ref{sec:ablation}) We evaluate the effect of progressively adding more training data from different datasets, including ScanNet++~\citep{yeshwanth2023scannet++}, RealEstate10K~\citep{zhou2018stereo}, and DL3DV~\citep{ling2024dl3dv}.}

\resizebox{\linewidth}{!}{
\begin{tabular}{l|l|cc|c|c}
\hline
\thickhline
\rowcolor{mygray} 
Exp ID & Datasets & mIoU$\uparrow$ & Acc.$\uparrow$ & PSNR$\uparrow$ & rel$\downarrow$ \\
\hline\hline
(1) & ScanNet & 0.5603 & 0.8297 & 25.48 & 3.18 \\ 
(2) & (1) + ScanNet++ & 0.5625 & 0.8334 & 25.65 & 3.10 \\ 
(3) & (2) + RE10K & 0.5717 & 0.8414 & 25.79 & 3.02 \\ 
(4) & (3) + DL3DV & \textbf{0.5755} & \textbf{0.8437} & \textbf{25.83} & \textbf{2.93} \\
\hline
\end{tabular}
}
\label{tab:ablation_data_scale}
\vspace{-3pt}
\end{table}

\begin{table}[!t]%[!ht]
% \vspace{-5pt}
\centering
\setlength{\abovecaptionskip}{2pt}
\caption{\textbf{Ablation on the number of input views.} (\S\ref{sec:ablation}) We fix the view range and evaluate under varying numbers of input views.}
\resizebox{1\linewidth}{!}{
\renewcommand{\arraystretch}{0.9}
\begin{tabular}{l|cc|cc|c}
\hline \thickhline
\rowcolor{mygray} 
\#Views & mIoU$\uparrow$ & Acc.$\uparrow$ & PSNR$\uparrow$ & SSIM$\uparrow$ & rel$\downarrow$ \\
\hline\hline
3 & 0.5574 & 0.8292  & 23.83 & 0.7683  & 4.12 \\
6 & 0.5846 & 0.8451  & 24.47 & 0.8138  & 3.79 \\
8 & 0.6027 & 0.8454  & 25.03 & 0.8652  & 3.30 \\
10 & 0.6227 & 0.8514 & 25.12 & 0.8723  & 3.21  \\
\hline
\end{tabular}
}
\label{tab:ablation_view_num}
\vspace{-3pt}
\end{table}

\begin{table}[!t]%[!ht]
\centering
\setlength{\abovecaptionskip}{2pt}
\caption{\textbf{Ablation on different mask Strategies.} (\S\ref{sec:ablation})}
\resizebox{1\linewidth}{!}{
\begin{tabular}{cc|cc|ccc}
\hline \thickhline
\rowcolor{mygray} Type of $\mathcal{M}_e$ & $\rho_e$ & Type of $\mathcal{M}_d$ & $\rho_d$ & mIoU$\uparrow$ & PSNR$\uparrow$ & rel$\downarrow$ \\
\hline \hline
Random & 0.50 &  N/A & 0 & 0.5412  & 24.64 & 3.45 \\
Croco  & 0.90 &  N/A & 0 & 0.5498  & 25.12 & 3.37 \\
Random & 0.50 &  3D GS & 0.30 & 0.5511  & 25.42 & 3.23 \\
\textbf{Random} & \textbf{0.50} &  \textbf{3D GS} & \textbf{0.50} & \textbf{0.5625}  & \textbf{25.65} & \textbf{3.10} \\
Random & 0.50 &  3D GS & 0.75 & 0.5470  & 25.10  & 3.38 \\
Random & 0.75 &  3D GS & 0.50 & 0.5284  & 24.55 &  3.56 \\
Random & 0.75 &  3D GS & 0.75 & 0.5231  & 24.21 &  3.62 \\
\hline
\end{tabular}
}
\label{tab:ablation_mask_ratio}
\vspace{-5pt}
\end{table}

\noindent\textbf{Ablation on Mask Strategy.} As shown in Table~\ref{tab:ablation_mask_ratio}, Croco masking~\citep{weinzaepfel2022croco} surpasses random masking under similar settings but remains weaker than our geometry-aware two-stage masking strategy. For random masking, overly low $\rho_e$ yields insufficient spatial reasoning, while excessively high $\rho_e$ hinders learning. The geometry-aware mask consistently improves both segmentation and reconstruction, indicating the balanced masking facilitates 3D representation learning.

%% file: sec/5_conclusion.tex
\section{Conclusion}

We introduced UniSplat, a feed-forward framework that learns unified 3D representations directly from unposed multi-view images. UniSplat addresses the key limitations of prior methods through three complementary components: a dual-masking strategy that strengthens geometry induction by enforcing structure reasoning from incomplete cues, a coarse-to-fine Gaussian splatting strategy that progressively refines scene appearance to capture both global structure and fine detail, and a pose-conditioned recalibration mechanism that enforces geometric–semantic consistency by aligning multi-head predictions in a shared  spatial frame. Together, these designs enable UniSplat to produce coherent, high-fidelity 3D representations that are robust to sparse-view settings and transferable across domains. Extensive experiments on both 3D vision benchmarks and embodied AI tasks confirm its effectiveness and versatility.

\small{\noindent\textbf{Acknowledgement.}
This work was supported by Zhejiang Provincial Natural Science Foundation of China (No. LR26F020002), National Natural Science Foundation of China (No. 62372405, 62472222, 62306292, U25A20442 and 62427808), Fundamental Research Funds for the Central Universities (226-2025-00057), the State Key Laboratory of Brain Cognition and Brain-inspired Intelligence Technology (No. SKLBI-K2025004), the Natural Science Foundation of Jiangsu Province (No. BK20240080), the National Defense Science and Technology Industry Bureau Technology Infrastructure Project (JSZL2024606C001), and Science and Technology Project of Jiangsu Geological Bureau (No. 2025KJ09).
}

%% file: sec/X_suppl.tex
\clearpage
\appendix
\maketitlesupplementary

\renewcommand\thefigure{A\arabic{figure}}
\setcounter{figure}{0}
\renewcommand\thetable{A\arabic{table}}
\setcounter{table}{0}
\renewcommand\theequation{A\arabic{equation}}
\setcounter{equation}{0}
\setcounter{footnote}{0}
\setcounter{section}{0}

\section*{Appendix}

The appendix is structured as follows:

\begin{itemize}
    \item \S\ref{sec:implementation_details} provides additional implementation details.
    \item \S\ref{sec:experimental_analysis} presents further experimental analyses. It includes:
    \begin{itemize}
        \item[*] \S\ref{sec:more_3d_results} provides supplementary results on 3D vision tasks.
        \item[*] \S\ref{sec:more_embodied_results} provides additional findings on embodied AI tasks.
        \item[*] \S\ref{sec:analysis_geometric_masking} examines geometric masking.
        \item[*] \S\ref{sec:more_ablation_study} reports extensive ablation studies.
        \item[*] \S\ref{sec:downstream_tasks} evaluates performance on downstream tasks.
    \end{itemize}
    \item \S\ref{sec:more_vis} supplements failure cases analysis and presents more qualitative results.
    \item \S\ref{sec:limitation_futurework} discusses the limitation and future work of our method.
\end{itemize}

\section{More Implementation Details}
\label{sec:implementation_details}

\subsection{3D Vision Tasks Implementation Details}
\label{sec:3d_tasks_details}
We adopt two complementary experimental settings for evaluating 3D vision tasks, each emphasizing different aspects of the unified 3D representation:

\noindent\textbf{LSM-based Setting.} Following ~\cite{fan2024large}, we train and evaluate on indoor scene datasets (ScanNet~\cite{dai2017scannet} and ScanNet++~\cite{yeshwanth2023scannet++}), where the model must jointly predicts geometry, appearance, and semantics from \emph{unposed} multi-view images.
This setting stresses \emph{semantic consistency} across views and the ability to lift 2D features into a coherent 3D semantic field.

\noindent\textbf{NoPoSplat-based Setting.} Following ~\cite{ye2025pose}, we train and evaluate on large-scale video datasets (RealEstate10K~\cite{zhou2018stereo} and ACID~\cite{liu2021infinite}) for \emph{pose-free} novel view synthesis.
This setting emphasizes \emph{geometric fidelity} and robustness to sparse, wide-baseline inputs without any pose supervision. The results are shown in ~\ref{sec:more_3d_results}.

\subsection{Embodied Tasks Implementation Details}
\label{sec:embodied_tasks_details}

To assess the effectiveness of UniSplat as a visual backbone for embodied AI, we evaluate our approach using both a synthetic reinforcement learning benchmark SPA~\cite{zhu2025spa} and a synthetic imitation learning benchmark RoboCasa~\cite{nasiriany2024robocasa}.

\subsubsection{Reinforcement Learning Benchmarks}

The RL-based benchmark~\cite{zhu2025spa} spans 268 tasks across 8 simulators and covers both single-task and language-conditioned multi-task scenarios. In all experiments, the UniSplat encoder is frozen and only the downstream policy network is trained, ensuring a fair comparison of representation quality.

\noindent\textbf{Single-task Benchmarks.} We include three representative single-task settings:
\begin{itemize}[leftmargin=*]
    \item \textbf{VC-1}~\cite{majumdar2023where}: 14 tasks from four simulators --- Adroit (AD)~\cite{kumar2016manipulators}, Meta-World (MW)~\cite{yu2020meta}, DMControl (DMC)~\cite{tunyasuvunakool2020dm_control}, and TriFinger (TF)~\cite{wuthrich2020trifinger}. Policies are 3-layer MLPs trained with 100 demonstrations per task (25 for MW) and evaluated over 50 rollouts using fixed seeds (100, 200, 300). The \texttt{[CLS]} token from UniSplat serves as the observation feature.
    \item \textbf{Franka Kitchen}~\cite{gupta2019relay}: a MuJoCo simulation of a Franka robot in a kitchen with 9-D joint-velocity action space. We evaluate five tasks: Sliding Door, Turn Light On, Open Door, Turn Knob, and Open Microwave. Policies are 2-layer MLPs trained on 25 demonstrations per task.
    \item \textbf{Meta-World~\cite{yu2020meta}}: 48 diverse manipulation tasks, encompassing easy, medium, and hard levels. We adopt the Diffusion Policy~\cite{chi2023diffusion} following~\cite{ze20243d}, training with 10 demonstrations and evaluating over 20 rollouts per task.
\end{itemize}

\noindent\textbf{Language-conditioned Multi-task Benchmarks.} We further evaluate our method on two multi-task benchmarks featuring natural-language instructions:
\begin{itemize}[leftmargin=*]
    \item \textbf{RLBench}~\cite{james2020rlbench}: 71 executable tasks split into two groups according to PolarNet~\cite{chen2023polarnet} categories (35 and 36 tasks). Each task has 100 demonstrations for training and 25 rollouts for evaluation. For each group, we train a language-conditioned multi-task agent. We use RVT-2~\cite{goyal2024rvt} as the policy backbone, replacing its convolutional encoder with our frozen UniSplat encoder.
    \item \textbf{LIBERO}~\cite{liu2023libero}: 130 tasks across five suites (Spatial, Object, Goal, LIBERO-10, LIBERO-90). We train the official transformer-based language-conditioned policy with 20 demonstrations per task, no data augmentation, and pre-extracted visual features from UniSplat.
\end{itemize}

\noindent\textbf{Policy Training and Evaluation.} For all settings, we adhere to the training hyperparameters and evaluation protocols of the respective benchmarks to ensure comparability with prior work~\cite{zhu2025spa}. The frozen UniSplat encoder outputs either the \texttt{[CLS]} token (for MLP/Diffusion policies) or unpatchified feature maps (for transformer-based policies).

\subsubsection{Imitation Learning Benchmarks}

UniSplat can extract informative visual representations from image observations, which should benefit imitation learning. We empirically validate this by evaluating UniSplat on the RoboCasa benchmark~\cite{nasiriany2024robocasa}. RoboCasa comprises 24 atomic kitchen tasks with associated natural-language instructions, spanning actions such as pick-and-place, opening, and closing.
For each task, we use a limited set of 50 human demonstrations and 3,000 synthetic demonstrations generated with MimicGen~\cite{mandlekarmimicgen}. We train our model on these demonstrations and use UniSplat as the state encoder. We benchmark against BC-Transformer~\cite{mandlekar2022matters} and the Gaussian World Model (GWM)~\cite{lu2025gwm}, reporting performance in terms of success rate.
The results are shown in \S\ref{sec:more_embodied_results}.

\section{More Experimental Analysis}
\label{sec:experimental_analysis}

\begin{table*}[!t]
    \footnotesize
    \centering
    \setlength{\abovecaptionskip}{2pt}
    \caption{\textbf{Quantitative comparison} of novel view synthesis on the RE10K dataset under various overlap settings (\S\ref{sec:more_3d_results}).}
    \resizebox{\textwidth}{!}{
    \begin{tabular}{l|ccc|ccc|ccc|ccc}
    \hline \thickhline
    \rowcolor{mygray} & \multicolumn{3}{c}{\textbf{Small}} & \multicolumn{3}{c}{\textbf{Medium}} & \multicolumn{3}{c}{\textbf{Large}} & \multicolumn{3}{c}{\textbf{Average}}\\ \cline{2-13}
    \rowcolor{mygray}\textbf{Method} & PSNR $\uparrow$ & SSIM $\uparrow$ & LPIPS $\downarrow$
    & PSNR $\uparrow$ & SSIM $\uparrow$ & LPIPS $\downarrow$
    & PSNR $\uparrow$ & SSIM $\uparrow$ & LPIPS $\downarrow$
    & PSNR $\uparrow$ & SSIM $\uparrow$ & LPIPS $\downarrow$ \\
    \hline\hline
    \multicolumn{13}{l}{\textit{Pose-Required}} \\
    pixelSplat & 20.277 & 0.719 & 0.265
    & 23.726 & 0.811 & 0.180
    & 27.152 & 0.880 & 0.121
    & 23.859 & 0.808 & 0.184  \\
    MVSplat & 20.371 & 0.725 & 0.250
    & 23.808 & 0.814 & 0.172
    & 27.466 & 0.885 & 0.115
    & 24.012 & 0.812 & 0.175  \\
    \hline
    \multicolumn{13}{l}{\textit{Supervised Pose-Free}} \\
    MASt3R & 16.305 & 0.516 & 0.451
    & 18.106 & 0.561 & 0.377
    & 17.975 & 0.524 & 0.402
    & 17.617 & 0.539 & 0.403 \\
    CoPoNeRF & 17.393 & 0.585 & 0.462
    & 18.813 & 0.616 & 0.392
    & 20.464 & 0.652 & 0.318
    & 18.938 & 0.619 & 0.388 \\
    Splatt3R & 17.789 & 0.582 & 0.375
    & 18.828 & 0.607 & 0.330
    & 19.243 & 0.593 & 0.317
    & 18.688 & 0.593 & 0.317 \\
    NoPoSplat & 22.514 & 0.784 & 0.210
    & 24.899 & 0.839 & 0.160
    & 27.411 & 0.883 & 0.119
    & 25.033 & 0.838 & 0.160 \\
    \hline
    \multicolumn{13}{l}{\textit{Self-Supervised Pose-Free}} \\
    SelfSplat & 14.828 & 0.543 & 0.469
    & 18.857 & 0.679 & 0.328
    & 23.338 & 0.798 & 0.208
    & 19.152 & 0.680 & 0.328 \\
    \textbf{Ours} & \textbf{22.765} & \textbf{0.789} & \textbf{0.205}
    & \textbf{25.246} & \textbf{0.845} & \textbf{0.156}
    & \textbf{27.872} & \textbf{0.891} & \textbf{0.113}
    & \textbf{25.397} & \textbf{0.843} & \textbf{0.157} \\
    \hline
    \end{tabular}
    }
    \label{tab:rek_results}
    \vspace{-5pt}
\end{table*}

\vspace{-3pt}
\subsection{More Results on 3D Vision Tasks}
\label{sec:more_3d_results}

\noindent\textbf{Novel View Synthesis.} As shown in Table~\ref{tab:rek_results}, we compare our method with several methods on the RE10K dataset across different overlap settings. Our method outperforms all previous pose-free methods and even surpasses some pose-required methods, demonstrating the effectiveness of our unified 3D representation and training strategy.
Qualitative results are shown in Fig.~\ref{fig:qualitative_nvs_re10k_small}-~\ref{fig:qualitative_nvs_re10k_large}.

\begin{table*}[!t]
\centering
\scriptsize
\setlength{\abovecaptionskip}{2pt}
\caption{\small \textbf{Multi-Task Imitation Learning Results in Robocasa} (\S\ref{sec:more_embodied_results}). Average success rates (\%) of multi-task agents trained with $50$ human demonstrations or $3000$ generated demonstrations per task. Results are evaluated over $50$ episodes.}
\setlength{\tabcolsep}{4pt}
\begin{tabular}{l|cccccccccccc}
\hline \thickhline
& \multicolumn{2}{c}{\makecell{\texttt{PnP}\\\texttt{CabToCounter}}}     & \multicolumn{2}{c}{\makecell{\texttt{PnP}\\\texttt{CounterToCab}}} &  \multicolumn{2}{c}{\makecell{\texttt{PnP}\\\texttt{CounterToMicrowave}}}   &
\multicolumn{2}{c}{\makecell{\texttt{PnP}\\\texttt{CounterToSink}}} &
\multicolumn{2}{c}{\makecell{\texttt{PnP}\\\texttt{CounterToStove}}}   &
\multicolumn{2}{c}{\makecell{\texttt{PnP}\\\texttt{MicrowaveToCounter}}}
\\
\cmidrule(lr){2-3} \cmidrule(lr){4-5} \cmidrule(lr){6-7} \cmidrule(lr){8-9} \cmidrule(lr){10-11} \cmidrule(lr){12-13}
\\[-13pt]                                                         \\
\vcell{Method}  & \vcell{H-50}  & \vcell{G-3000}  & \vcell{H-50}  & \vcell{G-3000}  & \vcell{H-50}  & \vcell{G-3000}  & \vcell{H-50}  & \vcell{G-3000}  & \vcell{H-50}  & \vcell{G-3000}  & \vcell{H-50}  & \vcell{G-3000} \\[-\rowheight]
\printcellbottom  & \printcellbottom & \printcellbottom  & \printcellbottom & \printcellbottom  & \printcellbottom & \printcellbottom   & \printcellbottom & \printcellbottom   & \printcellbottom & \printcellbottom   & \printcellbottom  & \printcellbottom  \\[0pt]
\midrule
BC-transformer & 2  & 18 & 6  & 28 & 2 & 18 & 2 & 44 & 2 & 6 & 2 & 8 \\
GWM  & 18  & 32 &  4 & 22 & 14 & 44 & 20 & 38 & 2 & 18 & 20 & 26 \\
Ours  & \textbf{25}  & \textbf{39} &  \textbf{10} & \textbf{34} & \textbf{17} & \textbf{51} & \textbf{26} & \textbf{47} & \textbf{8} & \textbf{22} & \textbf{29} & \textbf{36} \\
\midrule
& \multicolumn{2}{c}{\makecell{\texttt{PnP}\\\texttt{SinkToCounter}}} &
\multicolumn{2}{c}{\makecell{\texttt{PnP}\\\texttt{StoveToCounter}}} & \multicolumn{2}{c}{\makecell{\texttt{Open}\\\texttt{SingleDoor}}}      & \multicolumn{2}{c}{\makecell{\texttt{Open}\\\texttt{DoubleDoor}}} &
\multicolumn{2}{c}{\makecell{\texttt{Close}\\\texttt{DoubleDoor}}} &
\multicolumn{2}{c}{\makecell{\texttt{Close}\\\texttt{SingleDoor}}}
\\
\cmidrule(lr){2-3} \cmidrule(lr){4-5} \cmidrule(lr){6-7} \cmidrule(lr){8-9} \cmidrule(lr){10-11} \cmidrule(lr){12-13}
\\[-6pt]
\vcell{} & \vcell{H-50} & \vcell{G-3000} & \vcell{H-50} & \vcell{G-3000} & \vcell{H-50} & \vcell{G-3000} & \vcell{H-50} & \vcell{G-3000} & \vcell{H-50} & \vcell{G-3000} & \vcell{H-50} & \vcell{G-3000}  \\[-\rowheight]
\printcellbottom & \printcellbottom & \printcellbottom & \printcellbottom & \printcellbottom & \printcellbottom & \printcellbottom & \printcellbottom & \printcellbottom & \printcellbottom & \printcellbottom & \printcellbottom & \printcellbottom \\[1pt]
\midrule
BC-transformer & 8  & 42 & 6  & 28 & 46 & 50 & 28 & 48 & 28 & 46 & 56 & 94 \\
GWM  & 22  & 38 &  18 & 44 & 58 & 62 & 28 & 42 & 50 & 58 & 54 & 90 \\
Ours  & \textbf{29}  & \textbf{48} & \textbf{27}  & \textbf{46} & \textbf{63} & \textbf{68} & \textbf{34} & \textbf{51} & \textbf{54} & \textbf{64} & \textbf{60} & \textbf{96} \\
\midrule
& \multicolumn{2}{c}{\makecell{\texttt{Open}\\\texttt{Drawer}}} &
\multicolumn{2}{c}{\makecell{\texttt{Close}\\\texttt{Drawer}}} & \multicolumn{2}{c}{\makecell{\texttt{TurnOn}\\\texttt{Stove}}}      & \multicolumn{2}{c}{\makecell{\texttt{TurnOff}\\\texttt{Stove}}} &
\multicolumn{2}{c}{\makecell{\texttt{TurnOn}\\\texttt{SinkFaucet}}} &
\multicolumn{2}{c}{\makecell{\texttt{TurnOff}\\\texttt{SinkFaucet}}}
\\
\cmidrule(lr){2-3} \cmidrule(lr){4-5} \cmidrule(lr){6-7} \cmidrule(lr){8-9} \cmidrule(lr){10-11} \cmidrule(lr){12-13}
\\[-6pt]
\vcell{} & \vcell{H-50} & \vcell{G-3000} & \vcell{H-50} & \vcell{G-3000} & \vcell{H-50} & \vcell{G-3000} & \vcell{H-50} & \vcell{G-3000} & \vcell{H-50} & \vcell{G-3000} & \vcell{H-50} & \vcell{G-3000}  \\[-\rowheight]
\printcellbottom & \printcellbottom & \printcellbottom & \printcellbottom & \printcellbottom & \printcellbottom & \printcellbottom & \printcellbottom & \printcellbottom & \printcellbottom & \printcellbottom & \printcellbottom & \printcellbottom \\[1pt]
\midrule
BC-transformer & 42  & 74 & 80  & 96 & 32 & 46 & 4 & 24 & 38 & 34 & 50 & 72 \\
GWM  & 56  & 90 & 80 & 90 & 46 & 80 & 22 & 40 & 52 & 48 & 44 & 66 \\
Ours  & \textbf{60}  & \textbf{93} & \textbf{82}  & \textbf{96} & \textbf{51} & \textbf{84} & \textbf{29} & \textbf{49} & \textbf{55} & \textbf{56} & \textbf{58} & \textbf{77} \\
\midrule
& \multicolumn{2}{c}{\makecell{\texttt{Turn}\\\texttt{SinkSpout}}} &
\multicolumn{2}{c}{\makecell{\texttt{CoffeePress}\\\texttt{Button}}} & \multicolumn{2}{c}{\makecell{\texttt{TurnOn}\\\texttt{Microwave}}}      & \multicolumn{2}{c}{\makecell{\texttt{TurnOff}\\\texttt{Microwave}}} &
\multicolumn{2}{c}{\makecell{\texttt{CoffeeServe}\\\texttt{Mug}}} &
\multicolumn{2}{c}{\makecell{\texttt{CoffeeSetup}\\\texttt{Mug}}}
\\
\cmidrule(lr){2-3} \cmidrule(lr){4-5} \cmidrule(lr){6-7} \cmidrule(lr){8-9} \cmidrule(lr){10-11} \cmidrule(lr){12-13}
\\[-6pt]
\vcell{} & \vcell{H-50} & \vcell{G-3000} & \vcell{H-50} & \vcell{G-3000} & \vcell{H-50} & \vcell{G-3000} & \vcell{H-50} & \vcell{G-3000} & \vcell{H-50} & \vcell{G-3000} & \vcell{H-50} & \vcell{G-3000}  \\[-\rowheight]
\printcellbottom & \printcellbottom & \printcellbottom & \printcellbottom & \printcellbottom & \printcellbottom & \printcellbottom & \printcellbottom & \printcellbottom & \printcellbottom & \printcellbottom & \printcellbottom & \printcellbottom \\[1pt]
\midrule
BC-transformer & 54  & 96 & 48  & 74 & 62 & 90 & 70 & 60 & 22 & 34 & 0 & 12 \\
GWM  & 72  & 90 & 76 & 90 & 64 & 84 & 70 & 54 & 36 & 50 & 16 & 28 \\
Ours  & \textbf{79}  & \textbf{96} & \textbf{82}  & \textbf{93} & \textbf{70 } & \textbf{92} & \textbf{74} & \textbf{68} & \textbf{44} & \textbf{60} & \textbf{25} & \textbf{36} \\
\bottomrule
\end{tabular}
\vspace{-0.3cm}
\label{tab:robocasa-results}
\end{table*}

\subsection{More Embodied AI Results}
\label{sec:more_embodied_results}

\noindent\textbf{Robocasa Multi-Task Imitation Learning.} We further evaluate our UniSplat representation on the Robocasa~\cite{zhu2025spa} multi-task imitation learning benchmark, which includes 24 manipulation tasks in a simulated kitchen environment. As shown in Table~\ref{tab:robocasa-results}, our method consistently outperforms prior visual backbones, demonstrating the effectiveness of UniSplat for spatial reasoning and task execution in embodied AI.

\subsection{Analysis of Geometric Masking}
\label{sec:analysis_geometric_masking}

\noindent\textbf{Pose Estimation with Gaussian-based PnP Pose and Pose Head.} To test whether the dual masking improves geometry reasoning rather than merely regularizing features, we evaluate relative pose estimation using (i) RANSAC-PnP with the predicted 3D Gaussian centers from different fields and (ii) learnable pose heads. As shown in Table~\ref{tab:pose_estimation_ablation}, we observe stable trends: appearance field $\mathcal{G}_{\text{app}}$ yields the most reliable correspondences, anchor field $\mathcal{G}_{\text{anchor}}$ follows, coarse geometry field $\mathcal{G}_{\text{geo}}$ lags; and the final pose $C_\text{final}$ consistently refines the coarse estimate $C_\text{coarse}$. Because PnP relies exclusively on physically consistent 2D--3D correspondences, gains here indicate improved cross-view geometric consistency and provide empirical support that the proposed masking mechanism enhances geometry awareness.

\begin{table}[!ht]
\footnotesize
\centering
\setlength{\abovecaptionskip}{2pt}
\vspace{-5pt}
\caption{\textbf{Relative Pose Estimation} with Gaussian-based PnP Poses and Pose Heads (\S\ref{sec:analysis_geometric_masking}). We evaluate AUC with various thresholds on RealEstate10K~\cite{zhou2018stereo} and ACID~\cite{liu2021infinite} datasets.}
\begin{tabular}{l|ccc|ccc}
    \hline\thickhline
    \rowcolor{mygray}
     & \multicolumn{3}{c}{\textbf{RE10K}} & \multicolumn{3}{c}{\textbf{ACID}}  \\
    \cline{2-4} \cline{5-7}
    \rowcolor{mygray} \textbf{Method} &  5$^\circ$ $\uparrow$ & 10$^\circ$ $\uparrow$ & 20$^\circ$ $\uparrow$ &  5$^\circ$ $\uparrow$ &  10$^\circ$ $\uparrow$ &  20$^\circ$ $\uparrow$  \\
    \hline \hline
    $\mathcal{G}_{\text{geo}}$  & 0.510 & 0.664 & 0.796 & 0.305 & 0.472 & 0.612 \\
    $\mathcal{G}_{\text{anchor}}$  & 0.578 & 0.714 & 0.822 & 0.336 & 0.495 & 0.648 \\
    $\mathcal{G}_{\text{app}}$  & 0.592 & 0.736 & 0.834 & 0.344 & 0.504 & 0.653 \\
    $C_\text{coarse}$  & 0.524 & 0.693 & 0.805 & 0.311 & 0.487 & 0.625 \\\hline
    $C_\text{final}$ & \textbf{0.607} & \textbf{0.748} & \textbf{0.842} & \textbf{0.354} & \textbf{0.516} & \textbf{0.661} \\
    \hline
    \end{tabular}
    \label{tab:pose_estimation_ablation}
    \vspace{-5pt}
\end{table}

\subsection{More Ablation Studies}
\label{sec:more_ablation_study}

\noindent\textbf{Ablation on the Number of Gaussian Latent Tokens.} Table~\ref{tab:ablation_token_num} shows that increasing the number of Gaussian latent tokens from 64 to 128 improves mIoU, accuracy, and perceptual quality. 256 tokens give the best overall metrics with marginal gains over 128, while 512 tokens slightly degrade performance, suggesting diminishing returns and possible overfitting. An intermediate token count balances reconstruction fidelity and segmentation accuracy.

\begin{table}[!ht]
\centering
\setlength{\abovecaptionskip}{2pt}
\caption{\textbf{Ablation on the Number of Gaussian Latent Tokens} (\S\ref{sec:more_ablation_study}). We evaluate the effect of varying the number of Gaussian latent tokens on reconstruction and segmentation performance.}
\vspace{-3pt}
\resizebox{1\linewidth}{!}{
\begin{tabular}{c|cc|ccc}
\hline \thickhline
\rowcolor{mygray}
\# $T_\text{coarse}$ & mIoU$\uparrow$ & Acc.$\uparrow$ & PSNR$\uparrow$ & SSIM$\uparrow$ & rel$\downarrow$ \\
\hline\hline
64 & 0.5469 & 0.8237 & 25.05 & 0.8602 & 3.35 \\
128 & 0.5587 & 0.8284  & 25.33 & 0.8684 & 3.23 \\
256 & \textbf{0.5625} & \textbf{0.8334} & \textbf{25.65} & \textbf{0.8782} & \textbf{3.10} \\
512 & 0.5617 & 0.8312  & 25.53 & 0.8756 & 3.11 \\
\hline
\end{tabular}
}
\label{tab:ablation_token_num}
\vspace{-5pt}
\end{table}

\subsection{Downstream Tasks}
\label{sec:downstream_tasks}

We further validate the utility of UniSplat as a general 3D visual backbone on EmbodiedScan~\cite{wang2024embodiedscan}, a large-scale, ego-centric multi-modal 3D perception benchmark with oriented 3D boxes, semantic occupancy, and language prompts. We follow the official data organization, view sampling, and metric protocols to evaluate three downstream tasks: multi-view 3D detection, multi-view semantic occupancy predictions, and multi-view 3D visual grounding.
To adapt UniSplat for specific tasks, we append lightweight task-specific heads to its pretrained multi-view transformer encoder.
For 3D object detection, we attach a 3D detection head predicting oriented boxes (center, size, rotation) from UniSplat's fused multi-view geometric-semantic features.
For semantic occupancy prediction, a 3D decoder is added, taking the voxelized dense features to predict semantic grids.
For 3D visual grounding, we equip the 3D decoder with a cross-modal fusion transformer that integrates encoded language features with the 3D scene representation, followed by a grounding head sharing the detection architecture.
This setup directly measures how well self-supervised 3D representations learned from unposed images generalize to complex indoor perception tasks without any depth supervision.

\noindent\textbf{Multi-view 3D Object Detection.} As shown in Table~\ref{tab:mv_3dod}, UniSplat with RGB-only inputs consistently surpasses camera-only baselines and even strong RGB-D systems on EmbodiedScan, indicating that its unified geometric--semantic representation yields reliable oriented box estimates without depth supervision. We provide qualitative results in Fig.~\ref{fig:qualitative_3d_object_detection}.

\begin{table}[!ht]
\vspace{-5pt}
\setlength{\abovecaptionskip}{2pt}
\centering
\caption{\textbf{Multi-view 3D object detection} (\S\ref{sec:downstream_tasks}).}
\renewcommand{\arraystretch}{1}
\resizebox{1\linewidth}{!}{
\begin{tabular}{l|c|cccc}
\hline \thickhline
\rowcolor{mygray}
Method & Input &AP$_{25}$ & AR$_{25}$ & AP$_{50}$ & AR$_{50}$ \\
\hline\hline
ImVoxelNet~\cite{rukhovich2022imvoxelnet}& RGB & 6.15 & 20.39 & 2.41  & 6.31 \\
VoteNet~\cite{qi2019deep} &Depth & 3.20 & 6.11  & 0.38  & 1.22 \\
FCAF3D~\cite{rukhovich2022fcaf3d}& Depth & 9.07 & 44.23 & 4.11 & 20.22 \\
EmbodiedScan~\cite{wang2024embodiedscan}& RGB-D & 16.85 & 51.07 & 9.77 & 28.21 \\\hline
Ours& RGB & \textbf{28.69} & \textbf{62.24} & \textbf{15.34} & \textbf{39.57} \\
\hline
\end{tabular}
}
\label{tab:mv_3dod}
\vspace{-5pt}
\end{table}

\noindent\textbf{Multi-view Semantic Occupancy Prediction.} As shown in Table~\ref{tab:mv_occ}, UniSplat delivers markedly better voxel-level semantics than prior RGB methods and is competitive with or exceeds RGB-D variants, reflecting dense, scene-consistent 3D priors learned from unposed images.

\begin{table}[!ht]
\vspace{-3pt}
\centering
\setlength{\abovecaptionskip}{2pt}
\renewcommand{\arraystretch}{0.8}
\caption{\textbf{Multi-view Semantic Occupancy Prediction} (\S\ref{sec:downstream_tasks}).}
\resizebox{0.7\linewidth}{!}{
\begin{tabular}{l|c|c}
\hline \thickhline
\rowcolor{mygray}
Method & Input & mIoU \\
\hline\hline
OccNet~\cite{tong2023scene} & RGB & 8.07 \\
SurroundOcc~\cite{wei2023surroundocc} & RGB & 9.10 \\
EmbodiedScan~\cite{wang2024embodiedscan} & RGB & 10.48 \\
EmbodiedScan~\cite{wang2024embodiedscan} & RGB-D & 19.97 \\\hline
Ours & RGB & \textbf{27.45} \\
\hline
\end{tabular}
}
\label{tab:mv_occ}
\vspace{-5pt}
\end{table}

\noindent\textbf{Multi-view 3D Visual Grounding.} As shown in Table~\ref{tab:mv_vg}, UniSplat shows clear improvements over RGB-D baselines across overall, easy, and hard settings, demonstrating robust cross-modal alignment and spatial grounding from images alone.
Results show UniSplat provides consistent gains over camera-only baselines across all tasks, highlighting its effectiveness as a unified RGB-only 3D backbone.

\begin{table}[!ht]
\vspace{-3pt}
\centering
\setlength{\abovecaptionskip}{2pt}
\caption{\textbf{Multi-view 3D Visual Grounding} (\S\ref{sec:downstream_tasks}).}
\resizebox{0.9\linewidth}{!}{
\begin{tabular}{l|c|ccc}
\hline \thickhline
\rowcolor{mygray}
Method &Input & Overall & Easy & Hard  \\
\hline\hline
ScanRefer~\cite{chen2020scanrefer}  & RGB-D & 12.85 & 13.78 & 9.12   \\
BUTD-DETR~\cite{jain2022bottom}  & RGB-D & 22.14 & 23.12 & 18.23  \\
L3Det~\cite{zhu2023object2scene} & RGB-D & 23.07 & 24.01 & 18.34  \\
EmbodiedScan~\cite{wang2024embodiedscan} & RGB-D & 25.72 & 27.11 & 20.12  \\ \hline
Ours & RGB & \textbf{36.88} & \textbf{38.13} & \textbf{31.42} \\
\hline
\end{tabular}
}
\label{tab:mv_vg}
\vspace{-5pt}
\end{table}

\section{More Visualizations}
\label{sec:more_vis}

\subsection{Failure Cases}
\label{sec:failure_cases}
\begin{figure}[!ht]
\vspace{-3pt}
\centering
\setlength{\abovecaptionskip}{2pt}
\includegraphics[width=0.47\textwidth]{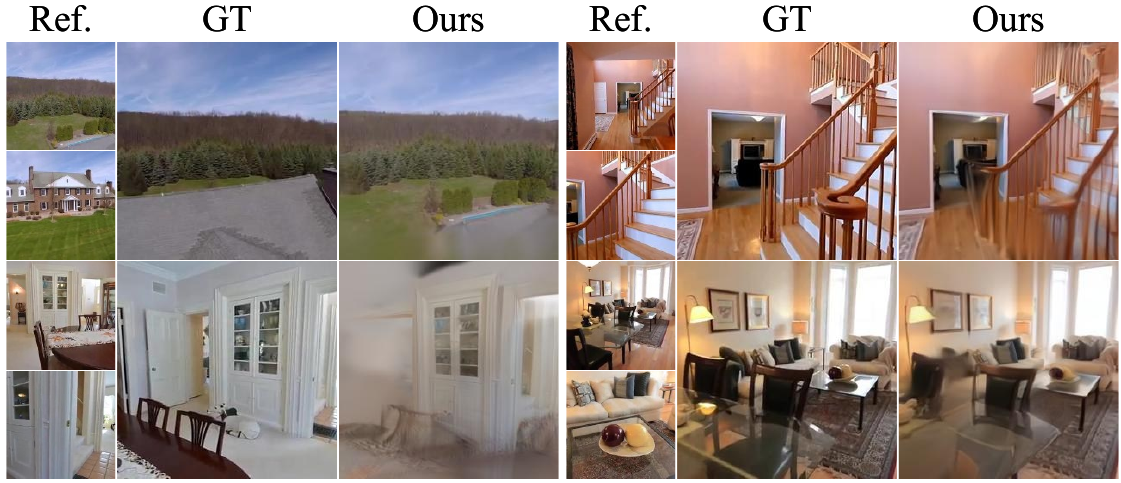}
\caption{\textbf{Failure cases} (\S\ref{sec:failure_cases}).}
\label{fig:failure_cases}
\vspace{-5pt}
\end{figure}

Fig.~\ref{fig:failure_cases} illustrates typical failure modes of our method, including blur and visual artifacts in occluded regions, as well as under extreme viewpoint changes.

\subsection{Novel View Synthesis Visualizations}
\label{sec:more_nvs_vis}
\begin{figure*}
\centering
\setlength{\abovecaptionskip}{2pt}
\includegraphics[width=0.8\textwidth]{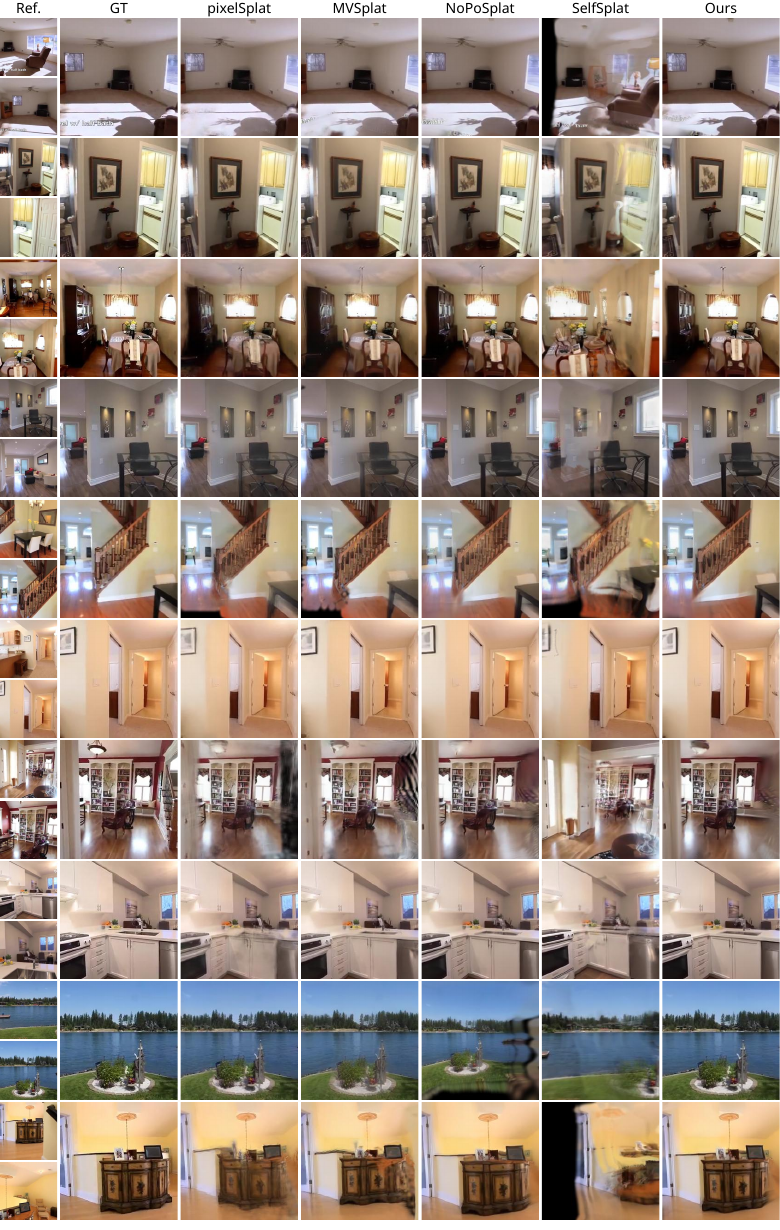}
\caption{\textbf{More qualitative comparisons on RE10K with small image overlap} (\S\ref{sec:more_nvs_vis}).}
\label{fig:qualitative_nvs_re10k_small}
\end{figure*}

\begin{figure*}
\centering
\setlength{\abovecaptionskip}{2pt}
\includegraphics[width=0.79\textwidth]{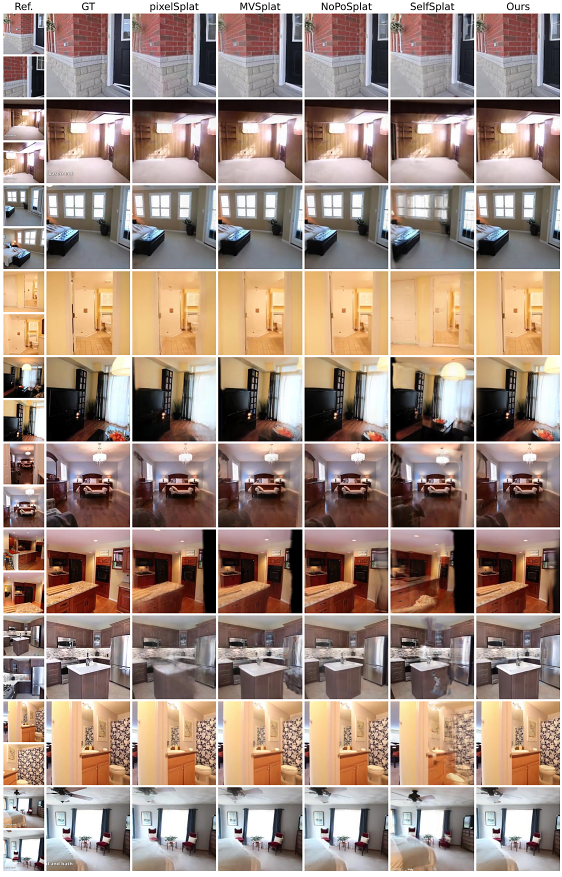}
\caption{\textbf{More qualitative comparisons on RE10K with medium image overlap} (\S\ref{sec:more_nvs_vis}).}
\label{fig:qualitative_nvs_re10k_medium}
\end{figure*}

\begin{figure*}
\centering
\setlength{\abovecaptionskip}{2pt}
\includegraphics[width=0.79\textwidth]{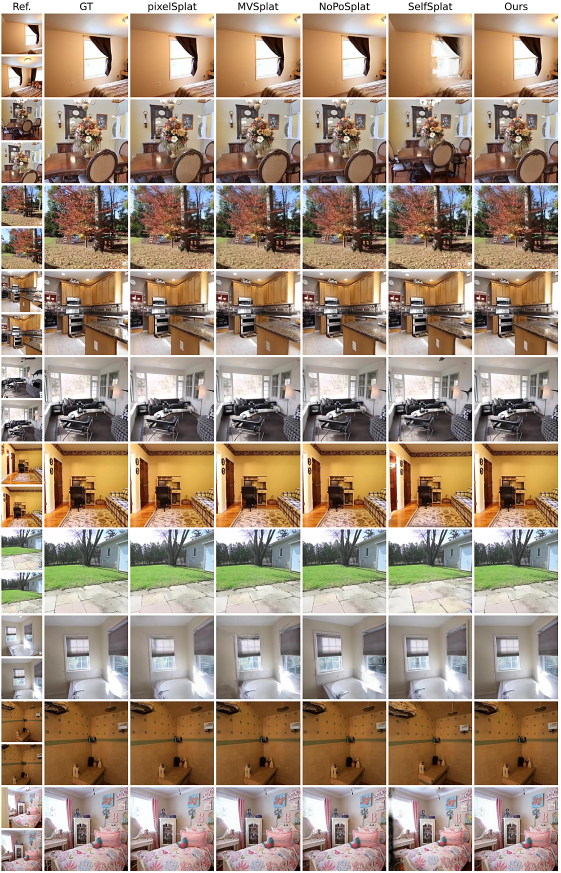}
\caption{\textbf{More qualitative comparisons on RE10K with large image overlap} (\S\ref{sec:more_nvs_vis}).}
\label{fig:qualitative_nvs_re10k_large}
\end{figure*}

\begin{figure*}
\centering
\setlength{\abovecaptionskip}{2pt}
\includegraphics[width=0.69\textwidth]{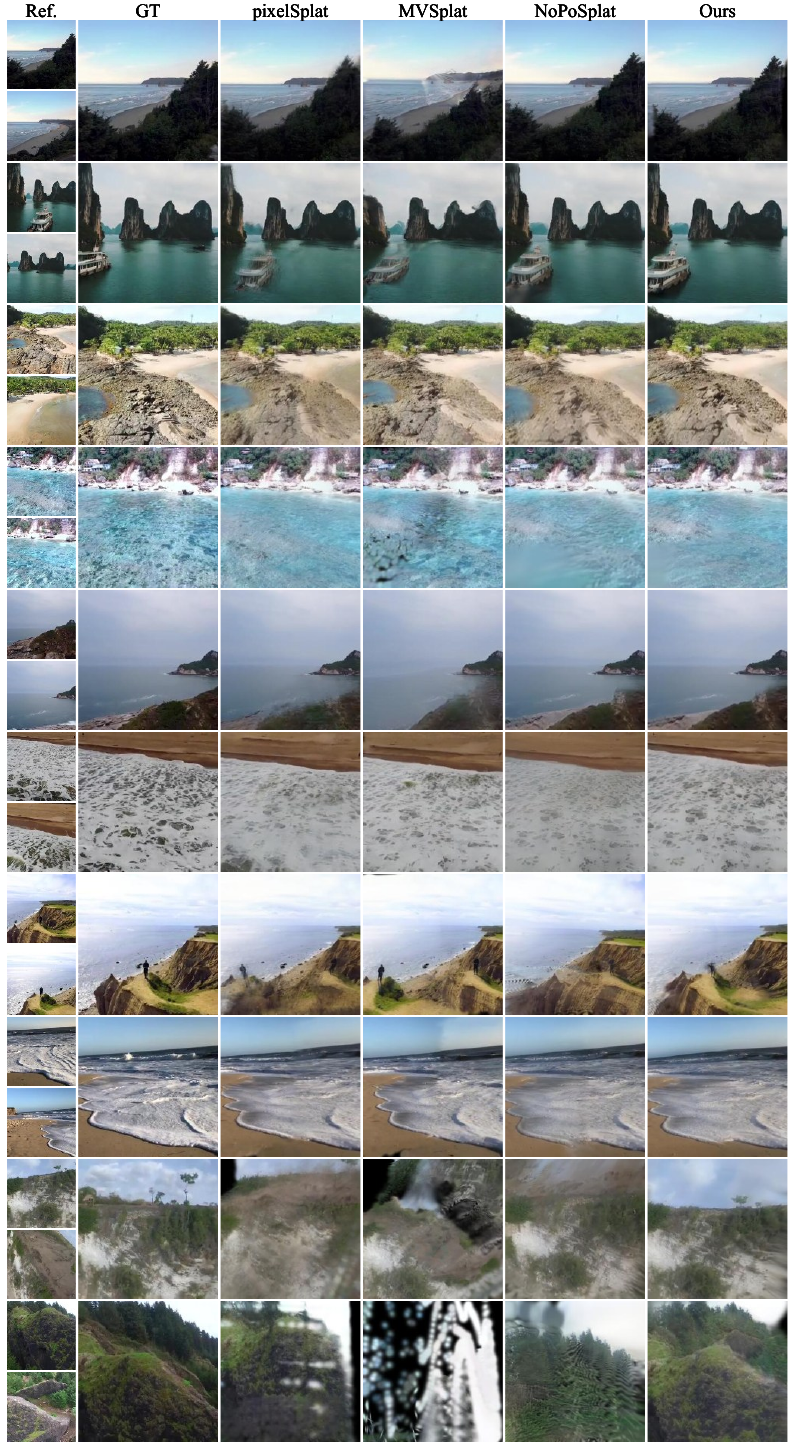}
\caption{\textbf{More qualitative comparisons on ACID} (\S\ref{sec:more_nvs_vis}).}
\label{fig:qualitative_nvs_acid}
\end{figure*}

We present additional qualitative results and visualizations to further illustrate the effectiveness of UniSplat across varying scene conditions and view overlaps, as shown in Fig.~\ref{fig:qualitative_nvs_re10k_small}--\ref{fig:qualitative_nvs_acid}.

\subsection{Semantic Segmentation Visualizations}
\label{sec:more_seg_vis}
\begin{figure*}[!t]
\centering
\setlength{\abovecaptionskip}{2pt}
\includegraphics[width=0.8\textwidth]{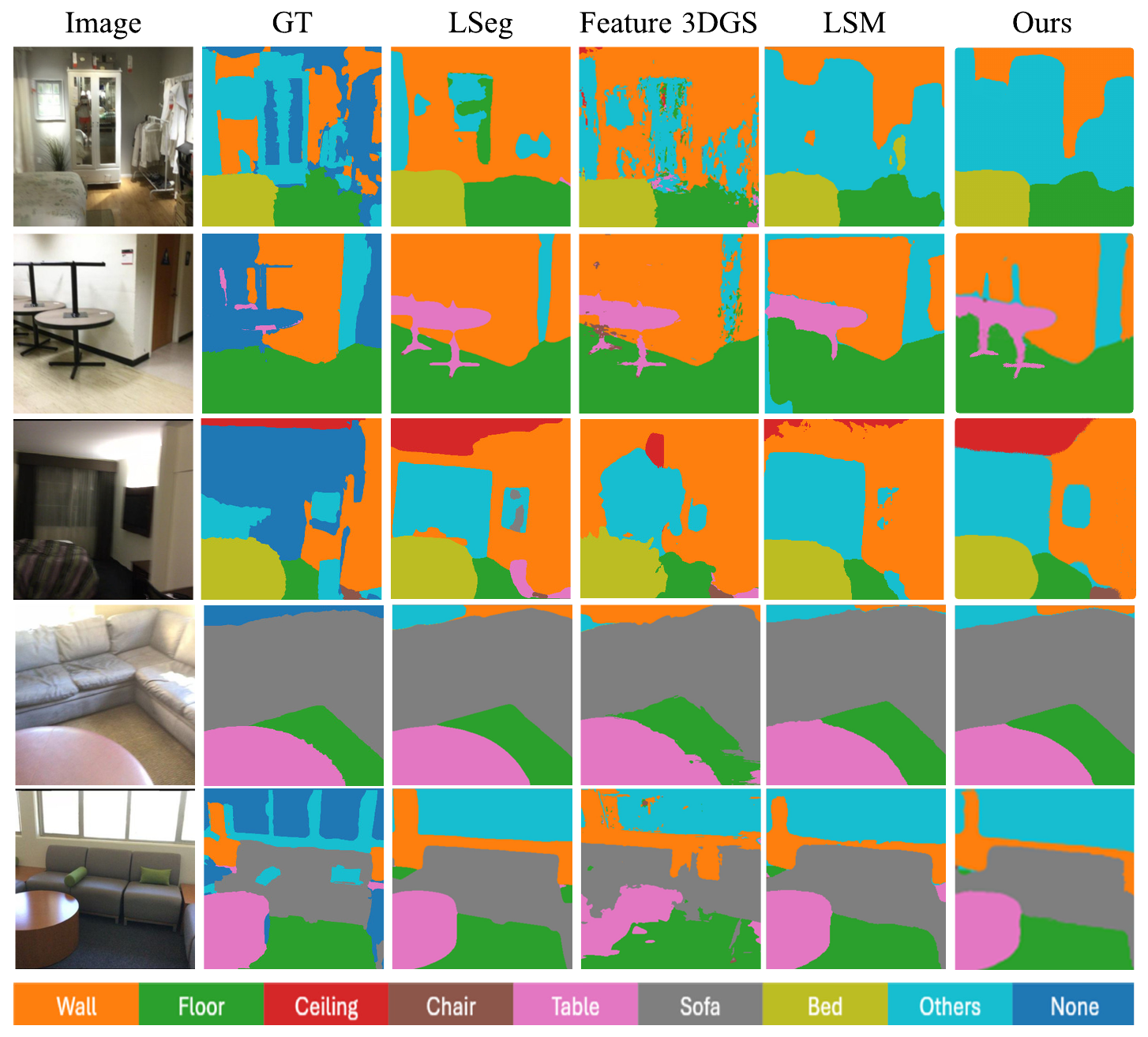}
\caption{\textbf{More qualitative comparison of novel-view segmentation on ScanNet} (\S\ref{sec:more_seg_vis}).}
\label{fig:qualitative_nvs_scannet_supp}
\vspace{10pt}
\end{figure*}

We provide more qualitative comparisons of novel-view semantic segmentation on ScanNet, as shown in Fig.~\ref{fig:qualitative_nvs_scannet_supp}.

\subsection{More Feature Map Visualizations}
\label{sec:more_fmap_vis}
\begin{figure*}[!t]
\centering
\vspace{10pt}
\setlength{\abovecaptionskip}{2pt}
\includegraphics[width=0.8\textwidth]{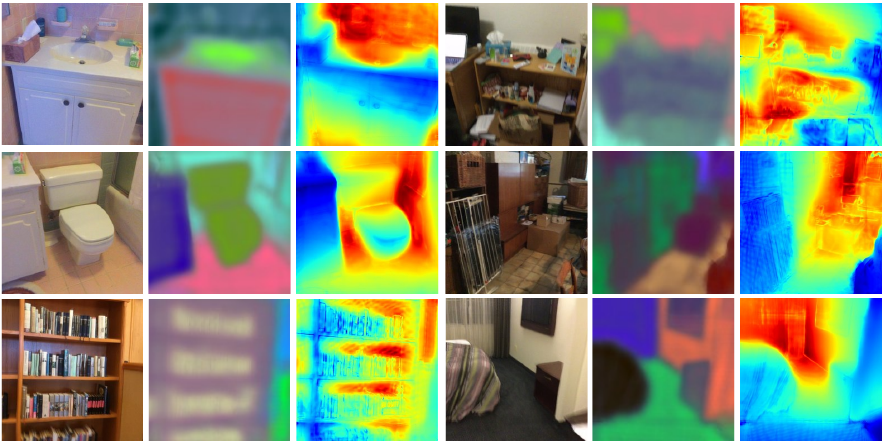}
\caption{\textbf{More visualizations of feature maps and depth maps} (\S\ref{sec:more_fmap_vis}).}
\label{fig:feature_map_vis}
\vspace{15pt}
\end{figure*}

We provide additional visualizations of the learned feature maps and depth maps from UniSplat in Fig.~\ref{fig:feature_map_vis}, demonstrating the model's ability to capture detailed geometric and semantic information.

\subsection{More 3D Object Detection Visualizations}
\label{sec:more_3d_object_detection_vis}
\begin{figure*}
\vspace{3cm}
\centering
\setlength{\abovecaptionskip}{2pt}
\includegraphics[width=0.8\textwidth]{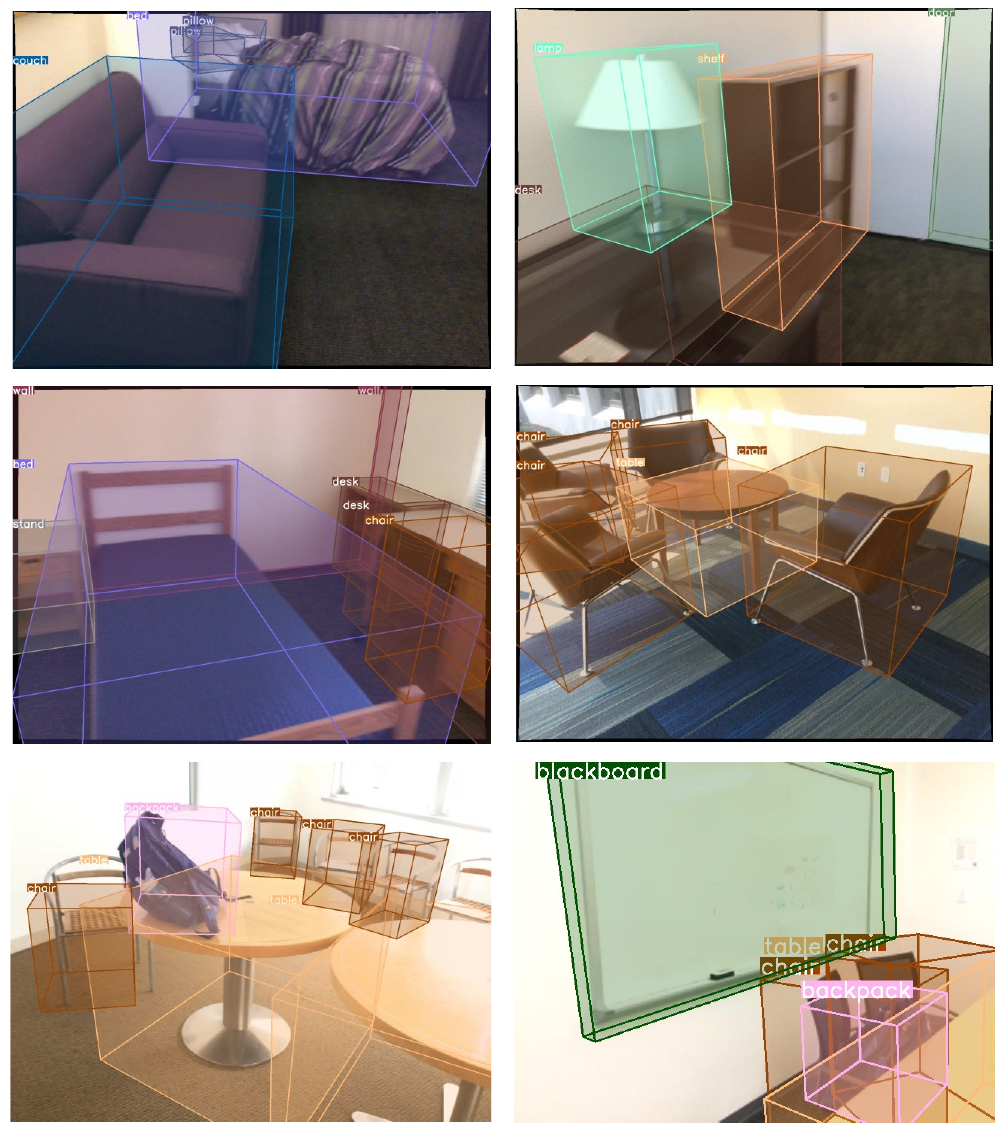}
\caption{\textbf{More qualitative results of 3D object detection} (\S\ref{sec:more_3d_object_detection_vis}).}
\label{fig:qualitative_3d_object_detection}
\vspace{3cm}
\end{figure*}

We include more qualitative results of 3D object detection on EmbodiedScan in Fig.~\ref{fig:qualitative_3d_object_detection}, showcasing the accurate oriented bounding box predictions achieved by UniSplat.

\section{Discussion and Limitations}
\label{sec:limitation_futurework}
\noindent\textbf{Limitations.}
Although UniSplat achieves strong performance across diverse 3D vision and embodied AI tasks, several limitations remain.
First, the framework still relies on pseudo-supervision from large pre-trained teacher models for geometry and semantics, which may propagate biases and inaccuracies from these teachers into the learned representation.
Second, while our geometry-aware masking and coarse-to-fine splatting improve robustness to sparse unposed views, performance degrades in extremely limited or highly textureless scenes, indicating that geometry induction could be further strengthened.
Third, our experiments are primarily conducted on indoor datasets; scaling to large-scale outdoor or highly dynamic environments may require additional adaptations, such as motion modeling or more robust pose estimation.
Finally, although the hierarchical Gaussian representation improves efficiency compared to dense splatting, rendering and training remain computationally intensive relative to purely latent approaches, which may limit deployment in resource-constrained settings.

\noindent\textbf{Future Work.}
Future research could address these limitations in several ways.
One direction is to develop geometry and semantic priors that are learned in a fully self-supervised manner, reducing dependence on external teacher models.
Another is to design adaptive masking and rendering strategies that adjust to scene complexity and viewpoint coverage, further improving robustness in sparse or degenerate cases.
Extending UniSplat to handle dynamic, open-world environments and outdoor scenes would broaden its applicability, potentially requiring integration of temporal modeling and more generalizable camera estimation.
Finally, incorporating large-scale language-scene interaction and multi-modal grounding could enable richer spatial reasoning and task understanding for embodied agents.

